\pdfoutput=1

\documentclass[11pt]{article}

\usepackage[final]{ACL2023}
\usepackage{times}
\usepackage{latexsym}
\usepackage{hyperref}
\usepackage{booktabs}
\usepackage{tabularx}
\usepackage{arydshln}
\usepackage{adjustbox}
\usepackage{tikz}
\usepackage{rotating}
\usepackage{multirow}
\usepackage{natbib}
\usepackage{multicol}
\usepackage{pgfplots}
\usepackage{graphicx}
\usepackage{amsmath}
\usepackage{pifont}
\usepackage{pdflscape}
\usepackage{fancybox}
\usepackage{pgfplotstable}
\pgfkeys{/pgf/number format/.cd,1000 sep={}}
\pgfplotsset{compat=1.17}
\usetikzlibrary{patterns}
\usepackage[normalem]{ulem}
\usepackage{svg}
\usepackage{enumitem}

\usepackage[T1]{fontenc}

\useunder{\uline}{\ul}{}

\usepackage[T1]{fontenc}

\usepackage[utf8]{inputenc}

\usepackage{microtype}

\usepackage{inconsolata}

\title{Controllable Text Summarization: Unraveling Challenges, Approaches, and Prospects - A Survey}

\author{Ashok Urlana\textsuperscript{1} \enspace \enspace Pruthwik Mishra\textsuperscript{2} \enspace \enspace Tathagato Roy\textsuperscript{2} \enspace \enspace Rahul Mishra\textsuperscript{2}\\
TCS Research, Hyderabad, India\textsuperscript{1} \enspace \enspace \enspace \enspace \enspace \enspace
IIIT Hyderabad\textsuperscript{2}\\
{\tt ashok.urlana@tcs.com}, {\tt pruthwik.mishra@research.iiit.ac.in}, \\{\tt tathagato.roy@research.iiit.ac.in}, {\tt rahul.mishra@iiit.ac.in}
}
\begin{document}
\maketitle
\begin{abstract}

Generic text summarization approaches often fail to address the specific intent and needs of individual users. Recently, scholarly attention has turned to the development of summarization methods that are more closely tailored and controlled to align with specific objectives and user needs. Despite a growing corpus of controllable summarization research, there is no comprehensive survey available that thoroughly explores the diverse controllable attributes employed in this context, delves into the associated challenges, and investigates the existing solutions. In this survey, we formalize the Controllable Text Summarization (CTS) task, categorize controllable attributes according to their shared characteristics and objectives, and present a thorough examination of existing datasets and methods within each category. Moreover, based on our findings, we uncover limitations and research gaps, while also exploring potential solutions and future directions for CTS. 
We release our detailed analysis of CTS papers at \url{https://github.com/ashokurlana/controllable\_text\_summarization\_survey}.
\end{abstract}
\section{Introduction} 
Despite the significant advancements in automatic text summarization, its one-size-fits-all approach falls short in meeting the varied needs of different segments of users and application scenarios. For example, generic automatic summarization may struggle to produce easily understandable summaries of scientific documents for non-expert users or create extremely brief summaries of news stories for online feeds. Lately, a myriad of works have emerged aimed at generating more controlled \cite{fan-etal-2018-controllable,maddela-etal-2022-entsum,he-etal-2022-ctrlsum,zhang2023macsum,pagnoni-etal-2023-socratic} and tailored text summaries that meet a wide range of user needs. 

\begin{figure}
    \centering
    \includegraphics[width=208pt]{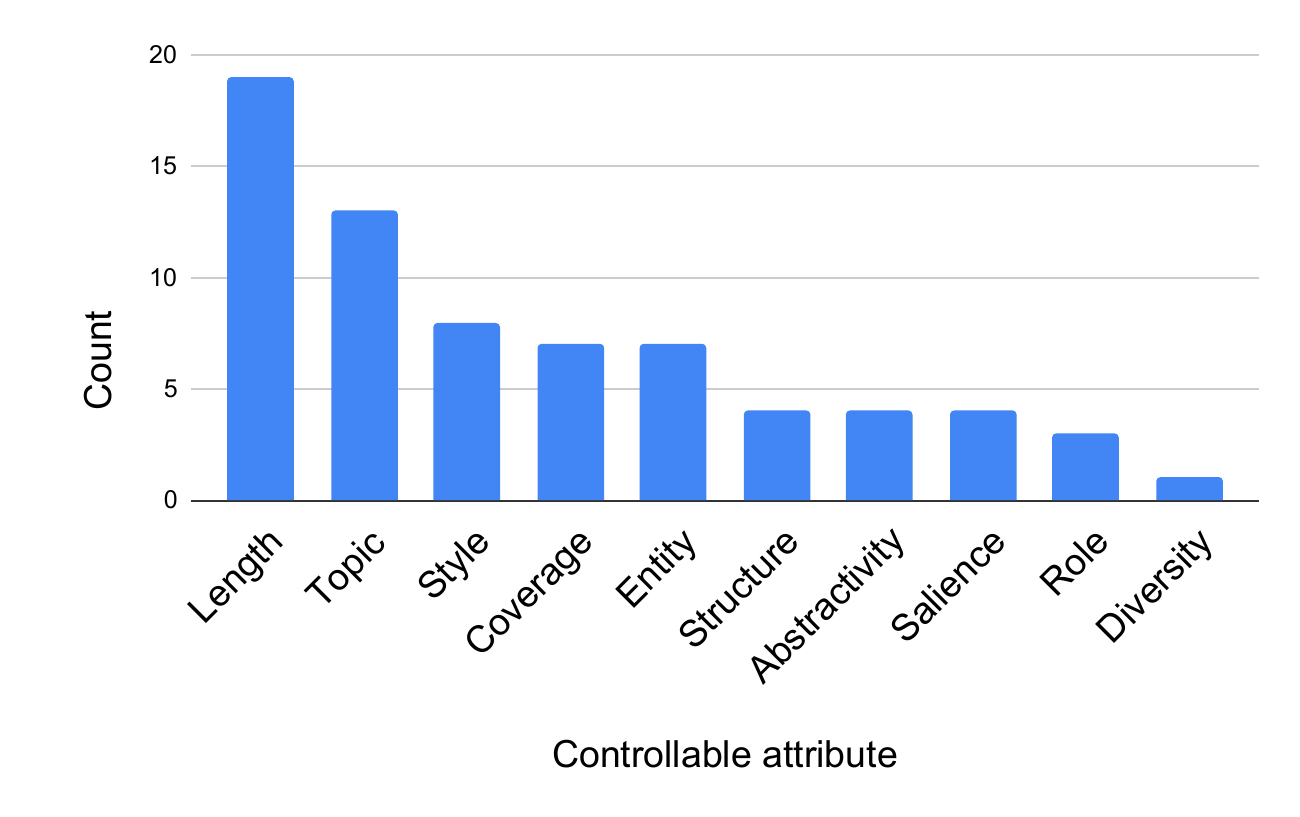}
    \caption{Number of controllable text summarization publications for various attributes.}
\vspace{-0.35cm}    
    \label{fig:count_control}
\vspace{-0.25cm}
\end{figure}

CTS task is centered around creating summaries of source documents that adhere to specific criteria. These criteria are managed through various controllable attributes (CA) or aspects like summary length \cite{kwon2023abstractive}, writing style \cite{goyal-etal-2022-hydrasum}, coverage of key information \cite{li-etal-2018-guiding, jin2020semsum}, content diversity \cite{narayan-etal-2022-well}, and more. These criteria vary depending on the task, user needs, and specific application context. For example, length-controlled summaries \cite{hitomi-etal-2019-large} are particularly useful in situations where brevity is crucial, like in social media posts, headlines, and abstracts. In areas such as marketing, academic writing, or professional communication, a style-controlled summary \cite{chawla-etal-2019-generating} is essential to ensure that the information aligns with the intended tone and messaging strategy. Similarly, topic-controlled summaries \cite{bahrainian2021cats} are commonly used in research papers, reports, and content curation, providing an emphasis on a specific topic to enhance clarity and coherence in the presented information. 

There is an uneven distribution of attention within the research community towards various CAs as depicted in Fig~\ref{fig:count_control}. The majority of CTS works concentrate on managing length, topic, and style. This could be attributed to two main factors. First, it is comparatively simpler to develop datasets for evaluating length, topic, and style compared to aspects like structure and diversity. Second, there is a plethora of application scenarios for length or topic-oriented summaries, such as generating concise news feeds or focused legal reports. 

In this survey, we collect and analyze 61 research papers pertaining to various possible CAs. The filtration criteria for the selection of papers are described in Appenedix~\ref{sec:papers_selection}. Subsequently, we classify these CAs into 10 categories, grouping similar ones based on shared characteristics and objectives. Moreover, we delve into the existing datasets, evaluating their creation methods and appropriateness for the respective task in each CA category. Furthermore, we scrutinize the current CTS methodologies for each CA category, drawing comparisons between their overarching frameworks and discussing relevant limitations. Subsequently, we discuss in detail the generic and specific evaluation strategies for CAs utilized by various works. Finally, we attempt to critique the current approaches and unravel potential future research trajectories. To the best of our knowledge, it is the first comprehensive survey on CTS.  
\begin{table}[t]
\centering\small
    \begin{tabularx}{\linewidth}{X}
    \toprule
    \textbf{Source:} (CNN)Novak \textbf{Djokovic} extended his current winning streak to 17 matches after beating Thomas Berdych 7-5, 4-6, 6-3 in the rain-interrupted \textbf{final} of the Monte Carlo Masters....After winning the Australian Open back in January, Djokovic has followed up with Masters' victories at Indian Wells and Miami. He then beat Rafa Nadal, arguably one of the greatest players on clay of all time... \\ \midrule
    \textbf{Length}:Long, \textbf{Coverage}:High, \textbf{Topic}:Djockovic,Final 
    \textbf{Summary:} Djokovic wins 7-5, 4-6, 6-3 after a tight match with Berdych in the Monte Carlo Masters final. Djokovic also followed up with Masters' victories at Indian Wells and Miami.\\ 
    \midrule
    \textbf{Length}:Normal, \textbf{Topic}:Djockovic, \textbf{Coverage}:Normal, 
    \textbf{Summary:} It's been a sensational year for Djokovic after beating Berdych in the finals and also winning against clay expert Nadal.\\ 
    \midrule
    \textbf{Length}:Short, \textbf{Coverage}:Normal, \textbf{Topic}: No Control\\
    \textbf{Summary:} Djokovic wins Monte Carlo Masters after beating Berdych 7-5, 4-6, 6-3 in the finals.   \\ \bottomrule       
    \end{tabularx}
    \caption{Summaries obtained by varying Controllable Attributes from MACSUM \cite{zhang2023macsum}}
    \label{tab:ex_macsum}
    \vspace{-0.25cm}
\end{table}


\section{Task Formulation}This section introduces the Controllable Text Summarization (CTS) task by outlining its definition and offering a categorized breakdown of different controllable attributes along with concise descriptions for each. Given a set of source documents $D = \{d_1, d_2, \ldots, d_k\}$. Each document, $d_i$, consists of a sequence of $n$ tokens: $\{x_{i,1}, x_{i,2}, \ldots, x_{i,n}\}$. $S_i$ is the target summary of document $d_i$, which comprises of a sequence of $m$ tokens: $\{s_{i,1}, s_{i,2}, \ldots, s_{i,m}\}$, where $m \ll n$. The user wants to control a set of controllable attributes $C$. The task can be framed as a conditional generative problem: 
$P(S|D, C) = \prod_{i}^{k} P(S_i|d_i, C)$

\subsection{Controllable Attributes}
A Controllable Attribute or Aspect (CA) refers to a user or application-driven trait of a summary designed to meet specific criteria or conditions, such as Length, Style, Role, etc. In the literature, it is evident that various authors use different terms to describe the same CAs, which exhibit similar characteristics and objectives (such as ``Salience: Key information", and ``Coverage: Granularity"). Additionally, numerous attributes can be encapsulated by a representative class; for instance, ``Style'' may serve as a class encompassing Tone, Readability, Humor, Romance, and similar aspects, facilitating their classification within the same category as shown in Table~\ref{tab:attrib_club_info}. Based on these observations, we group the CAs into 10 categories as listed in Table~\ref{tab:control_attributes}.
\begin{table}[t]
\centering\small
\renewcommand{\arraystretch}{0.9}
\setlength{\tabcolsep}{0.3ex}
\begin{tabularx}{0.49\textwidth}{lX}
\toprule
\multicolumn{1}{l}{\textbf{Attribute}} & \multicolumn{1}{c}{\textbf{Definition}} \\
\midrule
\textit{\textbf{Length}} & Controlling the length of the summary \\
\textit{\textbf{Style}} & Controlling the readability levels, politeness, humor, and emotion \\
\textit{\textbf{Coverage}} & Controlling the salient information in summary  \\
\textit{\textbf{Entity}} & Summary specific to pre-defined entities \\
\textit{\textbf{Structure}} &  Create summaries with predefined structure or order \\
\textit{\textbf{Abstractivity}} & Controlling the novelty in sentence formation \\
\textit{\textbf{Salience}} & Adjusting the presence of prominent information \\
\textit{\textbf{Role}} & Providing role-specific summaries \\
\textit{\textbf{Diversity}} & Generating semantically diverse summaries\\
\textit{\textbf{Topic}} & Controlling topic-focused summary generation\\
\bottomrule
\end{tabularx}
\caption{Controllable attributes definitions.}
\label{tab:control_attributes}
\vspace{-0.25cm}
\end{table}


\begin{table}[htb]
\centering
\begin{tabular}{ll}
\toprule
\textbf{Class}         & \textbf{Attribute}                                                                                                        \\ \midrule
Style         & \begin{tabular}[c]{@{}l@{}}Tone, Readability, Humor,
\\ Romance, Clickbait\end{tabular}                              \\ \midrule
Coverage      & Coverage, Granularity                                                                                            \\ \midrule
Entity        & Entity, Keyword                                                                                                  \\ \midrule
Topic         & \begin{tabular}[c]{@{}l@{}}Topic, Aspect, Decision of \\ interest, Opinion based on\\ user interest\end{tabular} \\ \midrule
Abstractivity & \begin{tabular}[c]{@{}l@{}}Abstractiveness, Extractiveness, \\ Novelty\end{tabular}                              \\ \midrule
Salience      & Salience, Key information                                                                                        \\ \bottomrule
\end{tabular}
\caption{Merging of attributes into representative classes.}
\label{tab:attrib_club_info}
\end{table}




\section{Related Surveys}
In the literature, a multitude of surveys center around conventional text summarization methods \citep{el2021automatic, nazari2019survey, allahyari2017text, gambhir2017recent}, including task-specific surveys such as multi-document summarization \citep{sekine-nobata-2003-survey}, cross-lingual summarization \cite{wang-etal-2022-survey}, and dialogue-based summarization \cite{tuggener-etal-2021-summarizing}. There are a few surveys that concentrate on text generation techniques \cite{zhang2023survey, prabhumoye-etal-2020-exploring} and the causal perspective \cite{wang2024recent, hu2021causal} on the same. On the contrary, this is the first survey, that focuses on controllable summarization by offering a thorough analysis of CTS methods, challenges, and prospects. 


\begin{table*}[htb]
\centering\scriptsize
\renewcommand{\arraystretch}{1.3}
\setlength{\tabcolsep}{1ex}
\begin{tabular}{lccllc}
\toprule
\textbf{Dataset} & \textbf{Controllable attribute(s)} & \textbf{Human-annotated} & \textbf{Size} & \textbf{Domain} & \textbf{Dataset URL} \\ \midrule
Multi-LexSum \citep{shen2022multi} & \phantom{0}\phantom{0}\phantom{0}Coverage & Yes & 9280 & Legal & \url{https://tinyurl.com/22ksfase} \\ 
GranDUC \citep{zhong-etal-2022-unsupervised}  & \phantom{0}\phantom{0}\phantom{0}Coverage & Yes & 50 & News & \url{https://tinyurl.com/2x72ubrw} \\
\midrule
TS and PLS \citep{luo-etal-2022-readability}  & Style & No & 28124 & Biomedical & \url{https://tinyurl.com/yck3v9px} \\
MACSUM \citep{zhang2023macsum} & Length, Coverage, Topic & Yes & 9686 & News, meetings & \url{https://tinyurl.com/3d2dsc7u} \\
\midrule
NEWTS \citep{bahrainian-etal-2022-newts} & Topic & Yes & 6000 & News & \url{https://tinyurl.com/36hzk3ew} \\ 
WikiAsp \citep{hayashi-etal-2021-wikiasp} & Topic & No & 320272 & Encyclopedia & \url{https://tinyurl.com/3u45hfbn} \\ 
ASPECTNEWS \citep{ahuja-etal-2022-aspectnews}  & Topic & No & 2000 & News & \url{https://tinyurl.com/bdzxs8ej} \\ 
Tourism ASPECTS \citep{mukherjee2020read}  & Topic & No & 7000 & Reviews & \url{https://tinyurl.com/ypjhhrxv} \\  
\midrule
EntSUM \citep{maddela-etal-2022-entsum} & \phantom{0}Entity & Yes & 2788 & News & \url{https://tinyurl.com/2pz9vzyw} \\
JAMUL \citep{hitomi-etal-2019-large}  & \phantom{0}\phantom{0}Length & No & 1932398 & News & \url{ https://tinyurl.com/3s3ecua9} \\ 
CSDS \citep{lin-etal-2021-csds}  & Role & Yes & 10700 & Dialogues & \url{ https://tinyurl.com/adk7zc7u} \\ 
MReD \citep{shen-etal-2022-mred}  & \phantom{0}\phantom{0}\phantom{0}Structure & Yes & 7089 & Meta reviews & \url{https://tinyurl.com/4nn87fd6} \\
\bottomrule
\end{tabular}%
\caption{List of controllable summarization datasets.}
\label{tab:dataset_info}
\vspace{-0.5cm}
\end{table*}
\section{Datasets}
\label{sec: datasets_info}
This section provides a broad overview of the CTS datasets and corresponding creation/acquisition strategies. The CTS methods are evaluated in several ways: 1) by utilizing publicly available summarization datasets, 2) by datasets derived from generic datasets, and 3) by creating human-annotated datasets. 
    \subsection{Generic Datasets}
    \label{sec:generic_datasets}
    CTS research predominantly leverages widely used news summarization datasets. Notably, about 57\% of CTS studies utilize either CNN-DailyMail \cite{nallapati-etal-2016-abstractive} or DUC \cite{over2004introduction, dang2005overview}. Other popular datasets, including Gigaword \cite{napoles2012annotated}, XSum \cite{narayan-etal-2018-dont}, NYTimes \cite{sandhaus2008new}, NEWSROOM \cite{narayan-etal-2018-dont}, and dialogue-based SAMSUM \cite{gliwa-etal-2019-samsum}, along with opinion-based datasets \cite{angelidis-lapata-2018-summarizing, angelidis-etal-2021-extractive}, are employed for controllable summarization. However, these generic datasets lack explicit annotations and nuances to evaluate the CA-specific summarization. CTS requires specialized datasets (as detailed in Table~\ref{tab:dataset_info}) to provide evaluation opportunities for specific aspects like length, topic, style, etc.
    \subsection{Derived Datasets}
    \label{sec:derived_datasets}
    The derived datasets are obtained by applying the aspect-specific heuristics to the widely used generic datasets. In this section, we list out a few derived datasets and their creation strategies.  \\
    \textbf{JAMUL.} \citet{hitomi-etal-2019-large} collect length-sensitive headlines for the Japanese language. Each article consists of three headlines with varying lengths of 10, 13, and 26 characters respectively. 
    \textbf{TS and PLS.} In order to enhance the readability of biomedical documents, \citet{luo-etal-2022-readability} introduce two types of summaries. The \textit{Technical Summary (TS)} is an abstract of a peer-reviewed bio-medical research paper and the \textit{Plain Language Summary (PLS)} is the authors submitted summary as part of the journal submission process. 
     \textbf{Wikiasp.} In order to construct the multi-domain aspect-based summarization corpus, \citet{hayashi-etal-2021-wikiasp} utilize the Wikipedia articles from 20 domains. Further, the section titles and paragraph boundaries of each article are obtained as a proxy of aspect annotation. In another study, \citet{ahuja-etal-2022-aspectnews} create the \textbf{ASPECTNEWS} dataset for aspect-oriented summarization. They achieve it by utilizing articles from the CNN/DailyMail dataset and identifying documents related to `earthquakes' and `fraud investigations' by using the universal sentence encoder \cite{cer2018universal}. Further, \citet{mukherjee2020read} collect a CA-based opinion summarization dataset consisting of tourism reviews. These are obtained from the \textit{TripAdvisor} website and identified the relevant aspects using the unsupervised attention-based aspect extraction technique \cite{he-etal-2017-unsupervised}.
    \subsection{Human annotated}
    \label{sec:human_annotated}
    This section provides the details of the human-annotated CTS datasets. \\
     \textbf{GranDUC.} By re-annotating the DUC-2004 \cite{dang2005overview}, \citet{zhong-etal-2022-unsupervised} release a novel benchmark dataset for the granularity control. Annotators are instructed to create summaries of multiple documents with \textit{coarse, medium,} and \textit{fine} granularity levels. 
   \textbf{Multi-LexSum.} \citet{shen2022multi} create a human-annotated corpus of 9,280 civil rights lawsuits and corresponding summaries with different degrees of granularity. 
The target summary length ranges from one-sentence to multi-paragraph level.
    %
    \textbf{EntSUM.} \cite{maddela-etal-2022-entsum} is a human-annotated entity-specific controllable summarization dataset. It utilizes the articles from The New York Times Annotated Corpus (NYT) \cite{sandhaus2008new} and includes annotated summaries for PERSON and ORGANISATION tags. The recent release of EntSUMV2 \cite{mehra-etal-2023-entsumv2} is the more abstractive version of EntSUM.
 \textbf{NEWTS.} \citet{bahrainian-etal-2022-newts} introduce the topically focused summarization corpus by leveraging documents from CNN-DailyMail and employing crowd-sourcing to generate two distinct summaries with different thematic aspects for each document.\\
\textbf{CSDS.} \citet{lin-etal-2021-csds} introduce the role-oriented Chinese Customer Service Dialogue Summarization (CSDS) dataset. It is meticulously annotated, segmenting the dialogues based on their topics and summarizing each segment as a QA pair.
 \textbf{MReD.} To tackle the task of structure-controllable summarization, \citet{shen-etal-2022-mred}  introduce the Meta-Review Dataset (MReD). It is created by gathering meta-reviews from the open review system and categorizing each sentence into one of nine predefined intent categories (abstract, strength, weakness, etc.,). 
 \textbf{MACSUM.} \citet{zhang2023macsum} develop a human-annotated corpus to control the mix of CAs (Topic, Speaker, Length, Extractiveness, and Specificity) together. MACSUM covers source articles from CNN/DailyMail and QMSUM \cite{zhong-etal-2021-qmsum} datasets. 
\section{Approaches to Controlled Summarization}
Various CAs have been investigated in controllable summary generation tasks, including style (politeness, humor, formality), content (length, entities, keywords), and structure. In this section, we describe various approaches to achieve CTS for the attributes mentioned in Table~\ref{tab:control_attributes}. Additionally, we list out the novel contributions and limitations for each paper in the Appendix~\ref{sec:models_desc} Table~\ref{tab:models_descritpion}.

\noindent \textbf{Length.} 
Earlier methods lacked length control 
and only employed heuristics such as stopping the generation after a fixed number of tokens.
To overcome this, four different approaches to integrate length as a learnable parameter are proposed. \\
\noindent \textbf{Adding length in input:} \citet{fan2018controllable} propose a convolutional encoder-decoder-based summarization system, where it quantizes summary lengths into discrete bins of different size ranges. During training, the input data is prepended with the gold summary length represented by bin lengths. Due to a fixed number of length bins, the system \underline{\textit{\textbf{fails}}} to generate summaries of arbitrary lengths. CTRLSUM \cite{he-etal-2022-ctrlsum} presents a generic framework to generate controlled summaries using keywords specific to length. 
Instead of controlling a single attribute, \citet{zhang2023macsum} allow different length attribute values (normal, short, long) to be used as inputs along with the source text for hard prompt tuning \cite{NEURIPS2020_1457c0d6}.

\noindent \textbf{Adding length in encoder:} \citet{yu-etal-2021-lenatten} propose a length context vector that is generated at each decoding step derived from the positional encodings. This vector is then concatenated with the decoder hidden state and encoder attention vectors. The 
\underline{\textit{\textbf{limitation}}} of the system is the generation of incomplete summaries for short desired lengths. \citet{liu-etal-2022-length} propose a length-aware attention model that adapts the source encodings based on the desired length by pretraining the model. \citet{zhang2023macsum} add a hyperparameter for learning the prefix embeddings for different attributes at each layer of the encoder and decoder for soft prefix tuning \cite{li-liang-2021-prefix}.

\noindent \textbf{Adding length in decoder:} \citet{kikuchi2016controlling} propose the first method to control length using a BiLSTM encoder-decoder architecture with attention \cite{luong2015effective} for sentence compression. In each step of the decoding process, an additional input for the remaining length is provided as an embedding. Instead of pre-defined length ranges, \citet{liu2018controlling} add a desired length parameter at the decoding step to each convolutional block of the initial layer of the convolutional encoder-decoder model. \citet{fevry-phang-2018-unsupervised} design an unsupervised denoising auto-encoder for sentence compression, where the decoder has an additional input of the remaining summary length at each time step. While it produces grammatically correct summaries, but they are nonsensical or semantically different from the input. This leads to the generation of \underline{\textit{\textbf{unfaithful}}} summaries.\\
To handle the length constraint, \citet{takase-okazaki-2019-positional} propose two modifications to the sinusoidal positional embeddings on the decoder side: length-difference positional encoding and length-ratio positional encoding. \citet{sarkhel-etal-2020-interpretable} present a multi-level summarizer that models a multi-headed attention mechanism using a series of 
interpretable semantic kernels to control lengths, reducing the trainable parameters significantly. The model does \underline{\textit{\textbf{not encode}}} the length attribute directly. \citet{song-etal-2021-new} design a confidence-driven generator that is trained on a denoising objective with a decoder-only architecture, where the source and summary tokens are masked with position-aware beam search. \citet{goyal-etal-2022-hydrasum} use a mixture-of-experts model with multiple transformer-based decoders for identifying different styles or features of summaries. \citet{kwon2023abstractive} introduce the summary length prediction task on the encoder side and this predicted summary length is inserted with a length-fusion positional encoding layer. \\
\textbf{Adding length in loss/reward function:} \citet{makino-etal-2019-global} propose a global minimum risk training optimization method under length constraint for the neural summarization tasks which is faster and generates five times fewer over-length summaries on an average than others. \citet{Chan2021ControllableSW} use an RL-based Constrained Markov Decision process with a mix of attributes. \citet{hyun-etal-2022-generating} devise an RL-based framework that incorporates both length and quality constraints in the reward function to generate multiple summaries of different lengths and according to the experimental results present in \citet{hyun-etal-2022-generating}, the model is computationally \underline{\textit{\textbf{expensive}}}.

\noindent \textbf{Style.} The generation of user-specific summaries has gained significant interest, but achieving distinct styles has posed an enduring challenge. These stylistic variations may encompass tone, readability control, or the modulation of user emotions. Style control aims to generate source-specific summaries \cite{fan-etal-2018-controllable} by utilizing the convolutional encoder-decoder network.\\ 
\citet{chawla-etal-2019-generating} obtain formality-tailored summaries by utilizing the input-dependent reward function. The pointer-generator \cite{see-etal-2017-get} network is used as the under-laying architecture and the loss function is modified with the addition of a formality-based-reward function. In another study, \citet{jin-etal-2020-hooks} attempt to control humor, romance, and clickbait in headlines using a multitask learning framework. 
   By employing an inference style classifier, \citet{cao-wang-2021-inference} adjust the decoder final states to obtain stylistic summaries. Moreover, they obtain lexical control by utilizing the word unit prediction that can directly constrain the output vocabulary. Similarly, \citet{goyal-etal-2022-hydrasum} extend the decoder architecture to a mixture-of-experts version by using multiple decoders. The gating mechanism helps to obtain multiple summaries for a single source. However, the major limitation in this model is its \underline{\textit{\textbf{manual}}} gating mechanism. To control various fine-grained reading grade levels, \citet{ribeiro2023generating} present three methods: instruction-prompting, reinforcement learning-based reward model, and look-ahead readability decoding approach.
    \\
    \noindent \textbf{Coverage.} Managing the information granularity is essential to measure the semantic coverage between the source text and the summary. To regulate the granularity, \citet{wu-etal-2021-controllable} introduce a two-stage approach, where the model incorporates a summary sketch, that encompasses user intentions and key phrases, serving as a form of weak supervision. They leverage a text-span-based conditional generation to govern the level of detail in the generated dialogue summaries. 
    \citet{zhong-etal-2022-unsupervised} propose a multi-granular event-aware summarization method composed of four stages: event identification, unsupervised event-based summarizer pretraining, event ranking, and summary generation by adding events as hints.
    Extraction of events from source text may \underline{\textit{\textbf{lower}}} the abstractiveness. \citet{zhang2023macsum} use the hard and soft-prompting strategies to control the amount of extracted text from the source in the summary. Additionally, \citet{huang-etal-2023-swing} utilize the natural language inference models to improve the coverage.
    
   \noindent \textbf{Entity.} Entity-centric summarization concentrates on producing a summary of a document that is specific to a given target entity \cite{hofmann-coyle-etal-2022-extractive}. \citet{zheng-etal-2020-controllable} extract the named entities using a pre-trained BERT \cite{devlin-etal-2019-bert} based model and feed both the article and the selected entities to a bidirectional LSTM \cite{hochreiter1997long} encoder-decoder model. In another study, \citet{liu-chen-2021-controllable} extract the entities (speakers and non-speaker entities) from a dialogue to form a planning sequence. The entities extracted 
   are concatenated to the source dialogue for training the conditional BART-based model. This model introduces factual \underline{\textit{\textbf{inconsistency}}} due to paraphrasing from a personal perspective.\\
  \citet{maddela-etal-2022-entsum} extend the GSum \cite{dou-etal-2021-gsum} 
  by feeding it either sentences or strings, which mention extracted entities as guidance. The model is an adapted version of BERTSum \cite{liu-lapata-2019-text}, where
  only the sentences containing the entity string mention and its coreferent mentions are fed. 
  \citet{hofmann-coyle-etal-2022-extractive} model entity-centric extractive summarization as a sentence selection task. Building upon BERTSum \cite{liu-lapata-2019-text}, they use a BERT \cite{devlin-etal-2019-bert} based encoders to represent the sentence and target entity pair and train with a contrastive loss objective to extract sentences most relevant to the target entities.\\
  \noindent \textbf{Structure.} Generic datasets lack key elements for emphasizing specific aspects in the corresponding ground truth summaries. To address this limitation and emphasize summary structure, \citet{shen-etal-2022-mred} achieve structure-controllable text generation by adding a control sequence at the beginning of the input text and treating summary generation as a standalone process. However, this approach has two main \underline{\textit{\textbf{limitations}}}, 1) generated tokens are solely based on logits predictions without ensuring that the sequence satisfies the control signal, 2) Auto-regressive models face error propagation in generation due to self-attention, causing subsequent generations to deviate from the desired output. To overcome these challenges, the sentence beam-search (SentBS) \cite{shen-etal-2022-sentbs} approach produces multiple sentence options for each sentence and selects the best sentence based on both the control structure and the model's likelihood probabilities. In a related study, \citet{zhong2023strong} utilize predicted argument role information to control the structure in legal opinion documents. Additionally, in the work of \citet{zhang2023macsum}, the prompt of entity chains, representing an ordered sequence of entities, is used for pre-training and fine-tuning with a planning objective to control the summary structure.

   \noindent \textbf{Abstractivity.} It measures the degree of textual novelty between the source text and summary. 
   \citet{see-etal-2017-get} introduce a pointer-generator network to control the source copying via \textit{pointing} and generate novel sentence formations by using \textit{generator} mechanism. However, this scheme \underline{\textit{\textbf{fails}}} to generate higher abstraction levels. \citet{kryscinski-etal-2018-improving} tackle this problem in two ways: 1) decompose the decoder into a contextual network to retrieve the relevant parts of the text and generate the summary by utilizing a pretrained model, 2) a mixed RL-based objective jointly optimizes the n-gram overlap with the ground truth summary. Similarly, \citet{song2020controlling} control the copying behavior by using a \textit{mix-and-match} strategy to generate summaries with varying n-gram copy rates. Based on the \textit{seen}, \textit{unseen} words from the source text, the system controls the copying percentage by acting as a language modeling task. Moreover, methods such as ControlSum \cite{fan-etal-2018-controllable} allow the users to explicitly specify the control attribute to facilitate better control. However, it does not provide any supervision on \underline{\textit{\textbf{violating}}} the controllability. To alleviate this issue, \citet{Chan2021ControllableSW} propose an RL-based framework on the constrained Markov decision process and introduced a reward to penalize the violation of attribute requirement.       
    %
    
    \noindent
    \textbf{Salience.} This attribute captures the most important information in a document. In SummaRuNNer \cite{nallapati2017summarunner}, salience is modeled as a feature in a classification objective. It uses GRU-based encoders and decoders to frame summarization as a text-to-binary sequence learning task at the sentence level \cite{bahdanau2014neural,cho-etal-2014-learning}. A binary score is assigned to each sentence, indicating its membership in the summary. The system performs \underline{\textit{\textbf{poorly}}} on out-of-domain datasets.
    To retain key content from the source, \citet{li-etal-2018-guiding} introduce a Key Information Guide Network, where keywords are identified by the TextRank algorithm with a modified attention mechanism that accommodates this key information as an additional input. However, it focuses mostly on informativeness \underline{\textit{\textbf{ignoring}}} coherence and readability features.\\
\citet{deutsch2023incorporating} model salience in terms of noun phrases using QA signals where the generation of the summary is conditioned on these identified phrases.
    This approach is \underline{\textit{\textbf{not applicable}}} to languages for which question generation and question answering models are not available. In long document CLS tasks, summarization systems often \underline{\textit{\textbf{fail}}} to respond to user queries. To resolve this issue, \citet{pagnoni-etal-2023-socratic} propose a pre-training approach that involves two tasks of salient information identification from sentences having the highest self ROUGE score and a question generation system to generate questions whose answers are the salient sentences. \\
    \textbf{Role.} Role-oriented dialogue summarization generates summaries for different roles/agents present in a dialogue (e.g. doctor and patient) \cite{liang-etal-2022-towards-modeling}. \citet{lin-etal-2021-csds} propose the CSDS dataset (see Section~\ref{sec:human_annotated}) and benchmark a variety of existing state-of-the-art summarization models for the task of generating agent and user surveys. They find that agent summaries generated by the existing methods \underline{\textit{\textbf{lack}}} key information, that needs to be extracted from dialogues of the other role.  To bridge this gap, \citet{lin-etal-2022-roles} build a role-aware summarization model for two users (agent and user) present in the dataset. They use two separate decoders for generating the user and agent summaries by utilizing user and agent masks. 
    A role attention mechanism is introduced to each decoder so that it can leverage the overall context by attending to the hidden states of the other role.
    \citet{liang-etal-2022-towards-modeling} use a role-aware centrality scoring model that computes role-aware centrality scores for each utterance, which measures the relevance between the utterance and the role prompts (signaling whether the summary is for the user or agent). This is then used to reweight the attention scores for each utterance, which is subsequently used by the decoder to generate the summary.\\
\textbf{Diversity.} Traditional decoding strategies, like beam search, excel at generating single summaries but often \underline{\textit{\textbf{struggle}}} to produce diverse ones. Techniques such as top-k and nucleus sampling are effective in generating diverse outputs but may sacrifice faithfulness. In response to these challenges, \citet{narayan-etal-2022-well} introduce compositional sampling, a decoding method to obtain diverse summaries. This method initiates by planning a semantic composition \cite{narayan-etal-2021-planning} of the target in the form of entity chains, and then leverages beam search to generate diverse summaries.\\
\noindent
\textbf{Topic.} Long documents often cover multiple topics, and a generic summary might not fully encompass the diverse scope. \citet{krishna-srinivasan-2018-generating} train a topic-conditioned pointer-generator network \cite{see-etal-2017-get} by concatenating one hot encoding representation of the topic with the embedding of each token in the input document. However, news categories are used as the predefined topics, that \underline{\textit{\textbf{limits}}} the generalization to other tasks. 
To handle diverse topics, \citet{tan-etal-2020-summarizing} utilize external knowledge sources like Wikipedia and ConceptNet to create a weakly supervised summarization framework compatible with any encoder-decoder architecture.
\citet{suhara-etal-2020-opiniondigest} propose an unsupervised method, where aspect-specific opinions are extracted from a set of reviews by a pre-trained opinion extractor, and the summary of the opinion is generated by a generator model trained to reconstruct the reviews from the opinions.
    Similarly, given a set of reviews for a product (e.g. Hotels), \citet{amplayo-etal-2021-aspect} train a Multiple Instance Learning (MIL) model, to extract the predictions for aspect (like cleanliness) codes at the document, sentence, token level \cite{mukherjee2020read}. These predicted aspects transform the input such that relevant sentences and keywords along with aspect tokens are fed into the pre-trained T5 \cite{2020t5} model.\\
\citet{hsu-tan-2021-decision} introduce the task of generating decision-supportive summaries. The focus is on predicting future Yelp ratings from the set of reviews using a Longformer-based \cite{Beltagy2020Longformer} regression model. They propose an iterative algorithm that selects the sentences of the summary from a set of representative sentences. 
    \citet{mukherjee-etal-2022-topic} extend topic-focused summarization for multimodal documents by creating a joint image-text context vector. 
\section{Evaluation Strategies}
This section catalogs and briefly describes the variety of automatic and human evaluation metrics that are being used to evaluate the summaries generated by the different methods studied in this paper. 
\subsection{Automatic Evaluation}
The automatic evaluation metrics can be categorized based on how they are defined. We categorize the metrics into n-gram-based, language-model-based, and aspect-specific.

\noindent \textbf{N-gram based} evaluation metrics like ROUGE \cite{lin-2004-rouge}, BLEU \cite{papineni-etal-2002-bleu} are based on matching n-grams from candidate summaries to a set of reference summaries. ROUGE is the most widely used metric in CTS literature. \textbf{Language-model based} metrics are computed using Pre-trained Language Models (PLM) like BERT \cite{devlin-etal-2019-bert} or BART \cite{Lewis2019BARTDS}. One class of approach computes the distance between the PLM embeddings of the reference and the generated summary.
Another way is based on computing the log probability of the generated text conditioned on input text as demonstrated in BARTScore \cite{NEURIPS2021_e4d2b6e6}. \textbf{Summarization specific} metrics including ROUGE-WE \cite{ng-abrecht-2015-better}, S\textsuperscript{3} \cite{peyrard-etal-2017-learning}, Sentence Mover's Similarity (SMS) \cite{clark-etal-2019-sentence}, SummQA \cite{scialom-etal-2019-answers}, BLANC \cite{vasilyev-etal-2020-fill}, and SUPERT \cite{gao-etal-2020-supert}, \textit{(Lite)\textsuperscript{3}Pyramid} \cite{zhang-bansal-2021-finding} are prominent for controllable summary evaluation.
\textbf{Aspect specific} metrics do not fall cleanly into either of the above-mentioned categories. These metrics focus on evaluating specific controllable aspects such as Flesh Reading Ease \cite{flesch1948}, Gunning Fog Index, and Coleman Liau Index for readability, control correlation, and error rate \cite{zhang2023macsum} for topic, abstractivity and role attributes. Appendix~\ref{sec:evaluation} Table~\ref{tab:automatic_evaluation_metrics} describes more details about the automatic evaluation metrics. 
\subsection{Human Evaluation}
Human evaluation is an indicator of the robustness and effectiveness of different summarization systems on specific aspects that cannot be directly captured by automatic evaluation metrics. These aspects include generic properties of a summary such as truthfulness \cite{song2020controlling,hyun-etal-2022-generating}, relevance \cite{goyal-etal-2022-hydrasum,he-etal-2022-ctrlsum,shen-etal-2022-mred}, fluency \cite{narayan-etal-2022-well,suhara-etal-2020-opiniondigest}, and readability \cite{cao-wang-2021-inference,kryscinski-etal-2018-improving} or specific properties such as completeness \cite{yu-etal-2021-lenatten,liu2022character} for length-controlled summaries, coverage \cite{mukherjee2020read,mukherjee-etal-2022-topic} for the entity, and topic-controlled summary generation. Broadly two kinds of scoring mechanisms are used for human evaluation: binary and rank-based. The rank-based scores usually range from 1 to 5. Despite these widely adapted mechanisms, human evaluation of summarization is challenging due to ambiguity and subjectivity. Aspects like coherence and fluency help mitigate ambiguity, but remain subjective to individual annotators. Accurately defining annotation descriptions is crucial, yet achieving a standardized approach across annotators remains difficult \cite{iskender-etal-2021-reliability, ito-etal-2023-challenges}.  The details about different human evaluation metrics are detailed in Appendix~\ref{sec:evaluation} Table~\ref{tab:human_evaluation}.

\section{Challenges and Future Prospects}
\textbf{Generic vs specialized benchmarks.} 
 We observe that more than 75\% of CTS works either utilize or alter the generic news summarization datasets to evaluate the controllable summarization. As shown in Table~\ref{tab:control_attributes}, out of the 10 categories, we could find CA-specific datasets for only seven categories. We envisage that conducting evaluations with specialized datasets that align closely with real-world application scenarios or user requirements will help better in assessing the practical utility, robustness, and performance of CTS. It is evident from our survey of CTS systems that evaluations are often confined to specific domains, like news, possibly due to the abundance of available datasets in that domain. However, this narrow focus limits the evaluation of the CTS model's robustness. 
\begin{table*}[ht]
\centering\small
\setlength{\tabcolsep}{0.7ex}
\begin{tabular}{lccccccc}
\toprule
\multirow{2}{*}{} & \multicolumn{7}{c}{\textbf{Controllable Attributes}} \\ \cmidrule(lr){4-6}
 & \multicolumn{1}{c}{\textbf{\textit{Length}}} & \multicolumn{1}{c}{\textbf{\textit{Entity}}} & \multicolumn{1}{c}{\textbf{\textit{Style}}} & \multicolumn{1}{c}{\textbf{\textit{Abstractivity}}} & \multicolumn{1}{c}{\textbf{\textit{Coverage}}} & \multicolumn{1}{c}{\textit{\textbf{Saliency}}} & \multicolumn{1}{c}{\textit{\textbf{Topic}}} \\ \midrule
\citet{fan-etal-2018-controllable} & \multicolumn{1}{c}{\ding{51}} & \multicolumn{1}{c}{\ding{51}} & \multicolumn{1}{c}{\ding{51}} & \multicolumn{1}{c}{\ding{55}} & \multicolumn{1}{c}{\ding{55}} & \multicolumn{1}{c}{\ding{55}} &  \multicolumn{1}{c}{\ding{55}}\\
\citet{zhang2023macsum} & \multicolumn{1}{c}{\ding{51}} & \multicolumn{1}{c}{\ding{55}} & \multicolumn{1}{c}{\ding{55}} & \multicolumn{1}{c}{\ding{55}} & \multicolumn{1}{c}{\ding{51}} & \multicolumn{1}{c}{\ding{55}} &  \multicolumn{1}{c}{\ding{51}}\\ 
\citet{Chan2021ControllableSW} & \multicolumn{1}{c}{\ding{51}} & \multicolumn{1}{c}{\ding{51}} & \multicolumn{1}{c}{\ding{55}} & \multicolumn{1}{c}{\ding{51}} & \multicolumn{1}{c}{\ding{55}} & \multicolumn{1}{c}{\ding{55}} &  \multicolumn{1}{c}{\ding{55}}\\ 
\citet{see-etal-2017-get} & \multicolumn{1}{c}{\ding{55}} & \multicolumn{1}{c}{\ding{55}} & \multicolumn{1}{c}{\ding{55}} & \multicolumn{1}{c}{\ding{51}} & \multicolumn{1}{c}{\ding{51}} & \multicolumn{1}{c}{\ding{55}} &  \multicolumn{1}{c}{\ding{55}}\\ 
\citet{pagnoni-etal-2023-socratic} & \multicolumn{1}{c}{\ding{55}} & \multicolumn{1}{c}{\ding{51}} & \multicolumn{1}{c}{\ding{55}} & \multicolumn{1}{c}{\ding{55}} & \multicolumn{1}{c}{\ding{55}} & \multicolumn{1}{c}{\ding{51}} &  \multicolumn{1}{c}{\ding{55}}\\ 
\citet{he-etal-2022-ctrlsum} & \multicolumn{1}{c}{\ding{51}} & \multicolumn{1}{c}{\ding{51}} & \multicolumn{1}{c}{\ding{55}} & \multicolumn{1}{c}{\ding{55}} & \multicolumn{1}{c}{\ding{55}} & \multicolumn{1}{c}{\ding{55}} & \multicolumn{1}{c}{\ding{55}} \\ 
\citet{nallapati2017summarunner} & \multicolumn{1}{c}{\ding{55}} & \multicolumn{1}{c}{\ding{55}} & \multicolumn{1}{c}{\ding{55}} & \multicolumn{1}{c}{\ding{51}} & \multicolumn{1}{c}{\ding{55}} & \multicolumn{1}{c}{\ding{51}} &  \multicolumn{1}{c}{\ding{55}}\\ \bottomrule
\end{tabular}%
\caption{Support for multiple controllable attributes across various models.}
\label{tab:multi_ca}
\end{table*}

\noindent \textbf{Standardization of metrics.} The goal of the CTS task is to produce CA-specific summaries, warranting a metric tailored to capture the nuances of this particular attribute. We observe that comparing models for a specific CA-based CTS task is challenging due to the use of varying metrics, leading each study to redo evaluations for a fair comparison with prior work. Standardizing CA-specific evaluation metrics could offer a valuable solution.\\
 \textbf{Explainability.} For effectively controlling user or application-specific attributes, it is imperative to leverage the understanding of the decision-making process within CTS systems. Also, this comprehension is essential for users or stakeholders, enabling them to discern how the system generates summaries from source text. This holds particular significance in applications where human decision-making or interpretation plays a pivotal role, such as in legal, medical, or financial domains. The existing CTS efforts lack proper emphasis on the explainability aspects, which can be readily addressed through the incorporation of suitable explanation methodologies \cite{abnar-zuidema-2020-quantifying, sundararajan2017axiomatic, lundberg2017unified}.


\noindent \textbf{Multi-lingual, multi-modal, and code-mixed CTS.} The existing literature on CTS predominantly focuses on works in English, with only one study addressing the topic in a Japanese context. We could not find any studies and datasets related to multilingual and code-mixed CTS approaches. Moreover, the task of controllable summarization in multi-modal and multi-document settings remains largely unexplored, presenting unique challenges for models to address and offering avenues for intriguing research problems.

\noindent \textbf{Multi-CA control.} Even though, few of the works perform multi-attribute controllable summarization \cite{goyal-etal-2022-hydrasum, he-etal-2022-ctrlsum, zhang2023macsum}, we observe that existing works predominantly investigate combinations of length and entity attributes (see Table~\ref{tab:multi_ca}). As a future research direction, it's essential to design models that consider other important combinations of control attributes, such as length, style, and saliency. Furthermore, creating standardized multi-CA benchmarks is crucial to facilitate the evaluations. 

\noindent \textbf{Reproducibility.} In the detailed analysis outlined in Table~\ref{tab:checklist}, we note that 35\% of research studies do not share their code publicly. Furthermore, 25\% of the papers did not carry out any human evaluation, and among the remaining studies, 79\% did not conduct Inter Annotator Agreement (IAA) assessments. The lack of reproducibility \cite{ito-etal-2023-challenges,gao-etal-2023-reproduction,iskender-etal-2021-reliability} measures hinders the scientific community's ability to validate and build upon existing work. On the other hand, the human study component should be a must for a text summarization evaluation scheme, otherwise, we are potentially overlooking essential aspects of real-world applicability. \\
\noindent \textbf{Standing on the shoulders of LLMs.} The rise and success of large language models (LLMs) have opened up unparalleled possibilities for leveraging their capabilities across diverse stages of the Natural Language Processing (NLP) pipeline. In the context of CTS, LLMs can be fine-tuned to grasp context-specific nuances about CAs without the need for a dedicated training set. Additionally, when it comes to evaluating CTS models, LLMs can serve as effective substitutes for human experts or judges (similar to \citet{liu2303g}), offering an 
efficient method for assessing performance. 



\section{Conclusions}
We present a comprehensive survey on controllable text summarization (CTS) by offering a detailed analysis, from formalizing various controllable attributes, classifying them based on shared characteristics, and delving into existing datasets, proposed models, associated limitations, and evaluation strategies. Moreover, we discuss the challenges and prospects, making it a helpful guide for researchers interested in CTS. We plan to keep the GitHub repository regularly updated with the latest CTS works. 
\section{Limitations}
Although we attempt to conduct a rigorous analysis of existing literature on controllable summarization, some works might have been possibly left out due to variations in search keywords. Furthermore, due to limited space, our survey primarily concentrates solely on the high-level aspects of the approaches, omitting a very fine-grained experimental comparison. Finally, our exploration of multilingual works was limited as we encountered challenges in finding them, likely influenced by the relatively low attention from the research community. We aim to further investigate the potential reasons behind the challenges associated with multilingual CTS tasks. 
\section{Ethics statement}
To uphold transparency and accountability, the papers utilized in this survey are detailed in Appendix~\ref{sec:papers_checklist} Table~\ref{tab:checklist}. We have provided a comprehensive set of papers, accompanied by our qualitative classification and annotations, enabling public scrutiny and examination. Moreover, to alleviate qualitative bias, each paper underwent review by at least three different individuals independently, aiming to minimize misclassification. We adhere to the same methodology to validate the presence of diverse observations in each paper. By incorporating these ethical considerations, we affirm our dedication to conducting research in an ethical and accountable manner. 


\bibliography{custom}
\bibliographystyle{acl_natbib}
\appendix

\section{Survey papers selection criteria}
\label{sec:papers_selection}
 We used keywords such as ``controllable summarization”, ``text summarization” and ``text generation” for selecting the initial pool of 105 papers. We selected the majority of papers from the reputed databases including the ACL Anthology\footnote{\url{https://aclanthology.org/}}, ACM Digital library\footnote{\url{https://dl.acm.org/}}, Google Scholar\footnote{\url{https://scholar.google.com/}}, which are known for hosting peer-reviewed articles that meet high academic standards. Among these 105 papers, six papers are pertinent to CTS, albeit they have not undergone peer review. Additionally, 23 papers touch upon the summarization aspect to some extent, although they may not be directly aligned with controllable summarization. Furthermore, we have excluded 15 papers as they primarily discuss controllable text generation or focus on enhancing the summarization task without specifically controlling any CTS attributes. Post to applying the above three filters we are left with 61 peer-reviewed and relevant papers to CTS. We have listed the filtration details in Table~\ref{tab:selction_papers}). 
 
\begin{table}[htb]
\centering
\begin{tabular}{lc}
\toprule
\textbf{Criteria} & \textbf{Number of papers} \\ \midrule
arXiv version & 6 \\ 
Not relevant & 23 \\ 
Enhancement & 15 \\ 
\textbf{Relevant} & \textbf{61} \\ \midrule
Total & 105 \\ \bottomrule
\end{tabular}
\caption{Survey papers filtration criteria.}
\label{tab:selction_papers}
\end{table}
\section{Evaluation Approaches}
\label{sec:evaluation}
We have listed the automatic and human evaluation methodologies along with their respective metric details in Table~\ref{tab:automatic_evaluation_metrics} and Table~\ref{tab:human_evaluation}. The automatic evaluation metrics are categorized into three groups: embedding-based, n-gram-based, and miscellaneous. Additionally, we present a compilation of papers organized by aspects, each associated with the relevant metrics, along with concise descriptions. As for human evaluation, we specify the corresponding metrics and provide definitions based on the attributes under consideration.
\begin{table*}[htb]
\centering\scriptsize
\renewcommand{\arraystretch}{1.3}
\setlength{\tabcolsep}{0.65ex}
\resizebox{\textwidth}{!}{
\begin{tabular}{cllll}
\toprule
\multicolumn{5}{c}{\textbf{Automatic Evaluation}} \\ \midrule
\multicolumn{1}{l}{\textbf{Type of metric}} & \multicolumn{1}{l}{\textbf{Attribute}} & \multicolumn{1}{l}{\textbf{Papers}} & \multicolumn{1}{l}{\textbf{Metrics}} & \textbf{Description} \\ \midrule
\multicolumn{1}{c}{\multirow{7}{*}{\begin{tabular}[c]{@{}l@{}} Embedding-based \\ (Language Model) \end{tabular}}} & \multicolumn{1}{l}{\multirow{4}{*}{General}} & \multicolumn{1}{l}{\citet{lin-etal-2022-roles}, \citet{liang-etal-2022-towards-modeling}} & \multicolumn{1}{l}{MoverScore} & \multirow{7}{*}{\begin{tabular}[c]{@{}l@{}}Computed using pretrained \\ language models, either by \\ computing similarity scores \\ between reference and \\ generated text embeddings \\ or through likelihood 
 \\ computation of the \\ generated text.\end{tabular}} \\ \cline{3-3} 
\multicolumn{1}{c}{} & \multicolumn{1}{c}{} & \multicolumn{1}{l}{\begin{tabular}[c]{@{}l@{}} \citet{song2020controlling}, \citet{shen-etal-2022-sentbs}, \\ \citet{cao-wang-2021-inference} , \citet{deutsch2023incorporating}, \\ \citet{Chan2021ControllableSW}, \citet{pagnoni-etal-2023-socratic}, \\  \citet{lin-etal-2022-roles}, \citet{liang-etal-2022-towards-modeling}, \\ \citet{narayan-etal-2022-well}, \citet{zhong2023strong}, \\ \citet{ribeiro2023generating}, \citet{shen2022multi}, \\ \citet{maddela-etal-2022-entsum}, \citet{lin-etal-2021-csds}\end{tabular}} & \multicolumn{1}{l}{BERTScore} &  \\ \cline{3-3}
\multicolumn{1}{c}{} & \multicolumn{1}{c}{} & \multicolumn{1}{l}{\citet{huang-etal-2023-swing}} & \multicolumn{1}{l}{BartScore} &  \\ \cline{3-3}
\multicolumn{1}{c}{} & \multicolumn{1}{c}{} & \multicolumn{1}{l}{\citet{zheng-etal-2020-controllable}} & \multicolumn{1}{l}{Bert-Reo} &  \\ \cline{3-3}
\multicolumn{1}{c}{} & \multicolumn{1}{l}{\multirow{3}{*}{Readability}} & \multicolumn{1}{l}{\multirow{3}{*}{\citet{luo-etal-2022-readability}}} & \multicolumn{1}{l}{\multirow{3}{*}{\begin{tabular}[c]{@{}l@{}}Masked Noun Phrase-based Text Complexity, \\ Ranked NP Based Text Complexity,\\ Masked Random Token-Based Text Complexity\end{tabular}}} &  \\
\multicolumn{1}{c}{} & \multicolumn{1}{c}{} & \multicolumn{1}{l}{} & \multicolumn{1}{l}{} &  \\
\multicolumn{1}{c}{} & \multicolumn{1}{c}{} & \multicolumn{1}{l}{} & \multicolumn{1}{l}{} &  \\ \midrule
\multicolumn{1}{l}{\multirow{4}{*}{Ngram Based}} & \multicolumn{1}{l}{\multirow{4}{*}{General}} & \multicolumn{1}{l}{\begin{tabular}[c]{@{}l@{}} \citet{lin-etal-2022-roles}, \cite{liang-etal-2022-towards-modeling}, \\ \citet{jin-etal-2020-hooks}, \citet{narayan-etal-2022-well} \end{tabular} } & \multicolumn{1}{l}{BLEU} & \multirow{4}{*}{\begin{tabular}[c]{@{}l@{}}These metrics are based on \\ matching ngram tokens \\ between reference and \\ generated summaries\end{tabular}} \\ \cline{3-3}
\multicolumn{1}{c}{} & \multicolumn{1}{c}{} & \multicolumn{1}{l}{\begin{tabular}[c]{@{}l@{}} \textbf{All except*} \citet{zhang2023macsum}, \citet{goldsack-etal-2023-biolaysumm}, \\ \citet{cao-wang-2021-inference}, \citet{hsu-tan-2021-decision}, \\ \citet{hofmann-coyle-etal-2022-extractive} \end{tabular}} & \multicolumn{1}{l}{ROUGE} &  \\ \cline{3-3}
\multicolumn{1}{c}{} & \multicolumn{1}{c}{} & \multicolumn{1}{l}{\citet{jin-etal-2020-hooks}, \citet{sarkhel-etal-2020-interpretable}} & \multicolumn{1}{l}{METEOR} &  \\ \cline{3-3}
\multicolumn{1}{c}{} & \multicolumn{1}{c}{} & \multicolumn{1}{l}{\citet{jin2020semsum}} & \multicolumn{1}{l}{Word Mover's Distance} &  \\ \midrule
\multicolumn{1}{l}{\multirow{14}{*}{Miscellaneous}} & \multicolumn{1}{l}{\multirow{5}{*}{Length}} & \multicolumn{1}{l}{\multirow{5}{*}{\begin{tabular}[c]{@{}l@{}} \citet{goyal-etal-2022-hydrasum}, \citet{kwon2023abstractive}, \\ \citet{liu2018controlling}, \citet{Chan2021ControllableSW} \end{tabular}}} & \multicolumn{1}{l}{\multirow{5}{*}{\begin{tabular}[c]{@{}l@{}}Absolute Length, Compression Ratio,\\ Length Variance, Var, Bin Percentage\end{tabular}}} & \multirow{14}{*}{\begin{tabular}[c]{@{}l@{}}Non-normative metrics proposed \\ by authors to evaluate specific \\ controlled aspect\end{tabular}} \\ 
\multicolumn{1}{c}{} & \multicolumn{1}{c}{} & \multicolumn{1}{l}{} & \multicolumn{1}{l}{} &  \\
\multicolumn{1}{c}{} & \multicolumn{1}{c}{} & \multicolumn{1}{l}{} & \multicolumn{1}{l}{} &  \\
\multicolumn{1}{c}{} & \multicolumn{1}{c}{} & \multicolumn{1}{l}{} & \multicolumn{1}{l}{} &  \\ \cline{3-3}
\multicolumn{1}{c}{} & \multicolumn{1}{l}{\multirow{2}{*}{Entity}} & \multicolumn{1}{l}{\citet{Chan2021ControllableSW}} & \multicolumn{1}{l}{QA-F1} &  \\ \cline{3-3}
\multicolumn{1}{c}{} & \multicolumn{1}{c}{} & \multicolumn{1}{l}{\citet{narayan-etal-2021-planning}} & \multicolumn{1}{l}{Entity Planning, Entity Specificity} &  \\ \cline{3-3} 
\multicolumn{1}{c}{} & \multicolumn{1}{l}{\begin{tabular}[c]{@{}l@{}}Topic, Speaker,\\ Length, Extractiveness, \\ Specificity\end{tabular}} & \multicolumn{1}{l}{\citet{zhang2023macsum}} & \multicolumn{1}{l}{\begin{tabular}[c]{@{}l@{}}Control Correlation,\\ Control Error Rate\end{tabular}} &  \\ \cline{3-3}
\multicolumn{1}{c}{} & \multicolumn{1}{l}{\multirow{2}{*}{\begin{tabular}[c]{@{}l@{}} Abstractiveness, \\ Degree of Specificity \end{tabular} }} & \multicolumn{1}{l}{\multirow{2}{*}{\citet{goyal-etal-2022-hydrasum}}} & \multicolumn{1}{l}{\multirow{2}{*}{Abstractiveness, Degree of specificity}} &  \\ 
\multicolumn{1}{c}{} & \multicolumn{1}{c}{} & \multicolumn{1}{l}{} & \multicolumn{1}{l}{} &  \\ \cline{3-3}
\multicolumn{1}{c}{} & \multicolumn{1}{l}{\multirow{4}{*}{Readability}} & \multicolumn{1}{l}{\citet{goyal-etal-2022-hydrasum}, \citet{cao-wang-2021-inference}} & \multicolumn{1}{l}{Dale-Chall} &  \\ \cline{3-3}
\multicolumn{1}{c}{} & \multicolumn{1}{c}{} & \multicolumn{1}{l}{\multirow{3}{*}{\citet{ribeiro2023generating}}} & \multicolumn{1}{l}{\multirow{3}{*}{\begin{tabular}[c]{@{}l@{}}Flesch Reading Ease, Gunning Fog Index,\\ Coleman Liau Index\end{tabular}}} &  \\
\multicolumn{1}{c}{} & \multicolumn{1}{c}{} & \multicolumn{1}{l}{} & \multicolumn{1}{l}{} &  \\
\multicolumn{1}{c}{} & \multicolumn{1}{c}{} & \multicolumn{1}{l}{} & \multicolumn{1}{l}{} &  \\ \bottomrule
\end{tabular}
}
\caption{Automatic evaluation metrics for controllable summarization, ``General'' refers to all controllable attributes.}
\label{tab:automatic_evaluation_metrics}
\end{table*}

\begin{table*}[htb]
\centering\small
\resizebox{\textwidth}{!}{
\begin{tabular}{llll}
\toprule
\multicolumn{4}{c}{\textbf{Human Evaluation}} \\ \midrule
\multicolumn{1}{c}{\textbf{Aspect(s)}} & \multicolumn{1}{c}{\textbf{Papers}} & \multicolumn{1}{l}{\textbf{Metrics}} & \textbf{Short description} \\ \midrule
\multicolumn{1}{l}{\begin{tabular}[c]{@{}l@{}}Abstractivity, Length, Style, \\ Topic, Coverage\\ Role, Entity\end{tabular}} & \multicolumn{1}{l}{\begin{tabular}[c]{@{}l@{}} \citet{sarkhel-etal-2020-interpretable}, \citet{song2020controlling} \\ \citet{liu-etal-2022-length}, \citet{cao-wang-2021-inference} \\ \citet{amplayo-etal-2021-aspect}, \citet{wu-etal-2021-controllable}, \\ \citet{jin2020semsum}, \citet{kwon2023abstractive}, \\ \citet{lin-etal-2022-roles}, \citet{liu-chen-2021-controllable}, \\ \citet{zheng-etal-2020-controllable}, \citet{bahrainian2021cats}, \\ \citet{suhara-etal-2020-opiniondigest}, \citet{lin-etal-2021-csds} \end{tabular}} & \multicolumn{1}{l}{Informativeness} & Has the summary covered key content of the input text? \\ \midrule
\multicolumn{1}{l}{\begin{tabular}[c]{@{}l@{}}Structure, Length, Entity, \\ Salience, Topic\end{tabular}} & \multicolumn{1}{l}{\citet{kwon2023abstractive}} & \multicolumn{1}{l}{Conciseness/Granularity} & Is the key information presented in a crisp way? \\ \midrule
\multicolumn{1}{l}{\begin{tabular}[c]{@{}l@{}}Structure, Style, Topic\\ Length,  Entity, Abstractivity, \\ Coverage, Role, Diversity\end{tabular}} & \multicolumn{1}{l}{\begin{tabular}[c]{@{}l@{}} \citet{goyal-etal-2022-hydrasum}, \citet{tan-etal-2020-summarizing} \\ 
\citet{yu-etal-2021-lenatten}, \citet{shen-etal-2022-mred}, \\ \citet{song2020controlling}, \citet{liu-etal-2022-length}, \\ \citet{fevry-phang-2018-unsupervised}, \citet{zheng-etal-2020-controllable}, \\ \citet{shen-etal-2022-sentbs}, \citet{cao-wang-2021-inference}, \\ \citet{amplayo-etal-2021-aspect}, \citet{hyun-etal-2022-generating}, \\ \citet{Chan2021ControllableSW}, \citet{jin2020semsum}, \\ \citet{liu2022character}, \citet{lin-etal-2022-roles}, \\ \citet{zhong-etal-2022-unsupervised}, \citet{jin-etal-2020-hooks}, \\ \citet{lin-etal-2021-csds} \end{tabular}} & \multicolumn{1}{l}{Fluency/Grammaticality} & Are the sentences in a summary grammatically correct? \\ \midrule
\multicolumn{1}{l}{Role, Topic, Diversity} & \multicolumn{1}{l}{\begin{tabular}[c]{@{}l@{}} \citet{narayan-etal-2022-well}, \citet{suhara-etal-2020-opiniondigest}, \\ \citet{lin-etal-2022-roles}, \citet{lin-etal-2021-csds}, \\ \citet{mukherjee2020read} \end{tabular}} & \multicolumn{1}{l}{Non-redundancy/Diversity} & Is the summary conveying diverse information? \\ \midrule
\multicolumn{1}{l}{Topic} & \multicolumn{1}{l}{\citet{krishna2018vocabulary}} & \multicolumn{1}{l}{Contextual Appropriateness} & Is the substituted word more readable in the summary? \\ \midrule
\multicolumn{1}{l}{Style, Diversity} & \multicolumn{1}{l}{\begin{tabular}[c]{@{}l@{}} \citet{goyal-etal-2022-hydrasum}, \citet{narayan-etal-2022-well}, \\ \citet{Chan2021ControllableSW}, \citet{jin2020semsum}, \\ \citet{zhong-etal-2022-unsupervised}, \citet{huang-etal-2023-swing} \end{tabular}} & \multicolumn{1}{l}{Faithfulness/Factuality} & \begin{tabular}[c]{@{}l@{}}Does the summary present factually correct content \\ with respect to the source?\end{tabular} \\ \midrule
\multicolumn{1}{l}{\begin{tabular}[c]{@{}l@{}}Style, Topic.\\ Entity, Structure\end{tabular}} & \multicolumn{1}{l}{\citet{goyal-etal-2022-hydrasum}, \citet{zhong2023strong}} & \multicolumn{1}{l}{Coherence} & Is the summary composed of correlated sentences? \\ \midrule
\multicolumn{1}{l}{\begin{tabular}[c]{@{}l@{}}Style, Structure, Length, \\ Entity, Abstractivity, \\ Coverage, Topic, Diversity\end{tabular}} & \multicolumn{1}{l}{\begin{tabular}[c]{@{}l@{}} \citet{goyal-etal-2022-hydrasum}, \citet{he-etal-2022-ctrlsum}, \\ \citet{shen-etal-2022-mred}, \citet{Chan2021ControllableSW}, \\ \citet{krishna-srinivasan-2018-generating}, \citet{shen-etal-2022-sentbs}, \\ \citet{cao-wang-2021-inference}, \citet{zhong-etal-2022-unsupervised}, \\ \citet{jin-etal-2020-hooks}, \citet{kryscinski-etal-2018-improving}, \\ \citet{luo-etal-2022-readability}, \citet{huang-etal-2023-swing} \end{tabular}} & \multicolumn{1}{l}{Relevance} & \begin{tabular}[c]{@{}l@{}}Does the summary contain relevant information \\ regarding the user provided attribute (topic/entity)?\end{tabular} \\ \midrule
\multicolumn{1}{l}{Abstractivity, Length} & \multicolumn{1}{l}{\begin{tabular}[c]{@{}l@{}} \citet{song2020controlling}, \citet{hyun-etal-2022-generating}, \\ \citet{fevry-phang-2018-unsupervised}, \citet{huang-etal-2023-swing} \end{tabular}} & \multicolumn{1}{l}{Truthfulness/Fidelity} & \begin{tabular}[c]{@{}l@{}}Has the summary successfully preserved the meaning \\ of the original text?\end{tabular} \\ 
\multicolumn{1}{l}{Length, Entity} & \multicolumn{1}{l}{\citet{he-etal-2022-ctrlsum}, \citet{yu-etal-2021-lenatten}} & \multicolumn{1}{l}{Accuracy/Correctness} & Is the information in the summary accurate? \\ \midrule
\multicolumn{1}{l}{Style} & \multicolumn{1}{l}{\begin{tabular}[c]{@{}l@{}}  \citet{kryscinski-etal-2018-improving}, \citet{ribeiro2023generating}, \\ \citet{cao-wang-2021-inference} \end{tabular}} & \multicolumn{1}{l}{Readability} & Is the text inside the summary readable? \\ \midrule 
\multicolumn{1}{l}{Length} & \multicolumn{1}{l}{\citet{yu-etal-2021-lenatten}, \citet{liu2022character}} & \multicolumn{1}{l}{Completeness} & Does the summary contain incomplete text? \\ \midrule
\multicolumn{1}{l}{Topic, Structure} & \multicolumn{1}{l}{\begin{tabular}[c]{@{}l@{}} \citet{zhong2023strong}, \citet{mukherjee2020read}, \\ \citet{mukherjee-etal-2022-topic}\end{tabular}} & \multicolumn{1}{l}{Coverage} & \begin{tabular}[c]{@{}l@{}}Does the summary include all the topics or aspects \\ defined in the source?\end{tabular} \\ \bottomrule
\end{tabular}
}
\caption{Human evaluation metrics for controllable text summarization.}
\label{tab:human_evaluation}
\end{table*}

\section{Model Descriptions}
As outlined in Table~\ref{tab:models_descritpion}, we augment novel contributions, utilized dataset, and the corresponding limitation for each paper, all aligned with the respective controllable attribute.
\label{sec:models_desc}
\begin{table*}[htb]
\centering\scriptsize
\setlength{\tabcolsep}{0.6ex}
\resizebox{\textwidth}{!}{%
\begin{tabular}{cllll}
\toprule
\textbf{Aspect} & \textbf{Paper} & \multicolumn{1}{c}{\textbf{Novel contribution}} & \textbf{Dataset(s)} & \textbf{Limitations} \\ \midrule
\multirow{4}{*}{Structure} & \citet{shen-etal-2022-mred} & Prepend structure prompt to the input & MRed & Subsequent generations deviate from the desired output \\ 
 & \citet{shen-etal-2022-sentbs} & Sentence-beam approach & MRed & Decoding methods significantly impact performance \\  
 & \citet{zhong2023strong} & Utilize predicted-role argument to control the structure & CanLII & Computationally expensive \\  \midrule
\multirow{4}{*}{Abstractivity} & \citet{see-etal-2017-get} & Pointer-generator network & CNNDM & \begin{tabular}[c]{@{}l@{}}Failed to achieve higher abstraction and \\ ineffective in core text selection\end{tabular} \\ 
 & \citet{kryscinski-etal-2018-improving} & \begin{tabular}[c]{@{}l@{}}Decouples the decoder into a contextual network and \\ mixed RL objective to encourage abstraction\end{tabular} & CNNDM & Less readable summaries \\ 
 & \citet{song2020controlling} & \begin{tabular}[c]{@{}l@{}}Mix-and-match strategy to generate summaries with \\ various degree of copying levels\end{tabular} & Gigaword, NEWSROOM & Poor performance in cross-domain settings \\ 
 & \citet{Chan2021ControllableSW} & \begin{tabular}[c]{@{}l@{}}RL-based framework on constrained markov decision \\ process to penalize the violation of control requirement\end{tabular} & CNNDM, NEWSROOM & Poor performance for highly abstractive targets \\ \midrule
Diversity & \citet{narayan-etal-2022-well}  & Compositional sampling decoding method & CNNDM, XSum & Generates unfaithful summaries for highly abstractive targets \\ \midrule
\multirow{7}{*}{Style} & \citet{fan2018controllable} & \begin{tabular}[c]{@{}l@{}}Convolutional encoder-decoder to generate stylistic \\ summaries by adding the source prompt to the input\end{tabular} & CNNDM, DUC2004 & Repetitive and longer summaries \\  
 & \citet{chawla-etal-2019-generating} & RL-based method to generate formality-tailored summaries & CNNDM, Webis-TLDR-17 & Poor performance in informal summary settings\\  
 & \citet{jin-etal-2020-hooks} & \begin{tabular}[c]{@{}l@{}}Multi-task learning framework with style-dependent \\ layer normalization and style-guided encoder attention\end{tabular} & \begin{tabular}[c]{@{}l@{}}NYT, CNN, Humor, \\ Romance, Clickbait corpus\end{tabular} & Poor performance on English Gigaword dataset \\
 & \citet{cao-wang-2021-inference} & \begin{tabular}[c]{@{}l@{}}Novel decoding methods: decode state adjustment, \\ word unit prediction based\end{tabular} & \begin{tabular}[c]{@{}l@{}}Hyperpartisan News \\ detection dataset\end{tabular} & - \\ 
 & \citet{goyal-etal-2022-hydrasum} & Mixture of experts strategy & CNNDM, XSum, NEWSROOM & Manual gating mechanism \\ 
 & \citet{luo-etal-2022-readability} & Readability control of bio-medical documents & LS, PLS & Fail to handle fine-grained readability control \\ 
 & \citet{ribeiro2023generating} & Fine-grained readability control & CNNDM & \begin{tabular}[c]{@{}l@{}}Style insights may not generalize beyond English \\ newswire datasets\end{tabular} \\ \midrule
\multirow{4}{*}{Coverage} & \citet{wu-etal-2021-controllable} & A two-stage control generation strategy & SAMSUM & - \\  
 & \citet{zhong-etal-2022-unsupervised} & \begin{tabular}[c]{@{}l@{}}Unsupervised framework to multi-granularity summary \\ generation\end{tabular} & \begin{tabular}[c]{@{}l@{}}Multi-NEWS, arXiv, \\ DUC2004\end{tabular} & \begin{tabular}[c]{@{}l@{}}Events extraction from source may effect the \\ abstractiveness\end{tabular} \\ 
 & \citet{huang-etal-2023-swing} & Utilize the NLI models to improve the coverage & DIALOGSUM, SAMSUM & Partially addressing the factuality problem \\ \midrule
\multirow{2}{*}{Role} & \citet{lin-etal-2022-roles} & \begin{tabular}[c]{@{}l@{}}Decoders for user and agent summaries and \\ attention divergence loss for the same topic\end{tabular} & CSDS, MC & - \\ 
 & \citet{liang-etal-2022-towards-modeling}  & \begin{tabular}[c]{@{}l@{}}Role aware centrality scores to reweight\\ context representations for decoding\end{tabular} & CSDS, MC & - \\ \midrule
\multirow{3}{*}{Entity} & \citet{zheng-etal-2020-controllable} & Controllable neural network with guiding entities & Gigaword, DUC 2004 & Performance poorer than SOTA models \\
 & \citet{liu-chen-2021-controllable} & \begin{tabular}[c]{@{}l@{}} Graph convolutional network based coreference fusion \\ layer and entity conditioned Summary Generation\end{tabular} & SAMSUM & \begin{tabular}[c]{@{}l@{}} Paraphrasing introduces factual inconsistencies \\ in person-specific summaries \end{tabular} \\
 & \citet{hofmann-coyle-etal-2022-extractive} & \begin{tabular}[c]{@{}l@{}}Model as a sentence selection task using transformer \\ based biencoder with a cosine similarity based loss\\ and adapting contrastive loss\end{tabular} & EntSUM & - \\ \midrule
\multirow{4}{*}{Salience} & \citet{nallapati2017summarunner}  & \begin{tabular}[c]{@{}l@{}}Summarization as a sentence selection task with salience \\ as a feature using sequence-to-sequence model\end{tabular} & CNNDM & Poor performance on out-of-domain datasets \\  
 & \citet{li-etal-2018-guiding} & Key information guided network with modified attention & CNNDM & Coverage mechanism not implemented \\
 & \citet{deutsch2023incorporating} & \begin{tabular}[c]{@{}l@{}}Model salience in terms of noun phrases by \\ incorporating QA signals\end{tabular} & CNNDM, DUC-2004 & \begin{tabular}[c]{@{}l@{}}Performance relies on question generation \\ and answering models\end{tabular} \\  
 & \citet{pagnoni-etal-2023-socratic} & \begin{tabular}[c]{@{}l@{}}Unsupervised pretraining involving \\ salient sentence selection\end{tabular} & QMSum, SQuALITY & Computationally expensive \\ \midrule
\multirow{19}{*}{Length} & \citet{kikuchi2016controlling} & Remaining words provided as additional input to decoder & Gigaword & Poor performance on DUC-2004 \\ 
 & \citet{fan2018controllable} & \begin{tabular}[c]{@{}l@{}}Convolutional encoder-decoder, summary length \\ grouping into bins and the source document \\ prepend with length bin's value\end{tabular} & CNNDM & Fails to generate summaries of arbitrary lengths \\ 
 &  \citet{liu2018controlling} & \begin{tabular}[c]{@{}l@{}}Remaining number of tokens replaced by characters at\\ the decoder\end{tabular} & CNNDM, DMQA & Fails to generalize to new control aspects at test time \\
 & \citet{fevry-phang-2018-unsupervised} & \begin{tabular}[c]{@{}l@{}}Unsupervised denoising auto-encoder for the task of \\ sentence compression and the decoder provided with \\ an additional input of the remaining summary length \\ at each time step\end{tabular} & Gigaword & Unfaithful summary generation in some cases \\ 
 & \citet{makino-etal-2019-global} & \begin{tabular}[c]{@{}l@{}}Global minimum risk training optimization method \\ under length constraint\end{tabular} & CNNDM, Mainichi & Fails to control length \\  
 & \citet{sarkhel-etal-2020-interpretable} & \begin{tabular}[c]{@{}l@{}}Multi-level summarizer with a multi-headed attention\\ mechanism using a series of timestep \\ independent semantic kernels\end{tabular} & \begin{tabular}[c]{@{}l@{}}MSR Narratives and \\ Thinking-Machines\end{tabular} & Fail to encode desired length \\ 
 & \citet{takase-okazaki-2019-positional} & \begin{tabular}[c]{@{}l@{}}Extension to the sinusoidal positional embeddings to \\ preserve the length constraint with length-difference \\ positional encoding and length-ratio positional encoding\end{tabular} & JAMUS corpus (Japanese) & Poor performance when desired target length is unseen \\  
 & \citet{yu-etal-2021-lenatten} & \begin{tabular}[c]{@{}l@{}}Concatenate the length context vector with the decoder \\ hidden state and other attention vectors\end{tabular} & CNNDM & Incomplete shorter summary generation \\  
 & \citet{song-etal-2021-new} & \begin{tabular}[c]{@{}l@{}}Confidence driven generator trained on a denoising \\ objective with a decoder only architecture with masked \\ source and summary tokens\end{tabular} & Gigaword, NEWSROOM & Poor performance on large datasets \\  
 & \citet{Chan2021ControllableSW} & \begin{tabular}[c]{@{}l@{}}Used a reinforcement learning based Constrained \\ Markov Decision Process to control length along \\ with constraints on a mix of attributes such \\ as abstractiveness and covered entity\end{tabular} & \begin{tabular}[c]{@{}l@{}}CNNDM, NEWSROOM \\ DUC-2002\end{tabular} & Length control only at word level \\
 & \citet{liu2022character} & \begin{tabular}[c]{@{}l@{}}Dynamic programming algorithm based on the \\ Connectionist Temporal Classification model\end{tabular} & Gigaword, DUC2004 & \begin{tabular}[c]{@{}l@{}}Poor performance compared to \\ autoregressive models\end{tabular} \\ 
 & \citet{goyal-etal-2022-hydrasum} & Mixture-of-expert model with multiple decoders & \begin{tabular}[c]{@{}l@{}}CNNDM, XSum, \\ NEWSROOM\end{tabular} & \begin{tabular}[c]{@{}l@{}}No insights about style diversity in \\ non-English and non-newswire datasets\end{tabular} \\ 
 & \citet{he-etal-2022-ctrlsum} & A generic framework using keywords & CNNDM, arXiv, BIGPATENT & High reliance on the quality of extracted keywords \\ 
 & \citet{liu-etal-2022-length} & \begin{tabular}[c]{@{}l@{}}Length aware attention model adapting the source \\ encodings\end{tabular} & CNNDM, XSum & \begin{tabular}[c]{@{}l@{}}Performance directly proportional to the \\ summary length\end{tabular} \\
 & \citet{zhong-etal-2022-unsupervised} & \begin{tabular}[c]{@{}l@{}}Events identification with unsupervised summary \\ generation\end{tabular} & \begin{tabular}[c]{@{}l@{}}GranuDUC, MultiNews, \\ DUC2004, arXiv\end{tabular} & Fails to capture abstractness due to event extraction \\ 
 & \citet{hyun-etal-2022-generating} & \begin{tabular}[c]{@{}l@{}}RL based framework incorporating both the length \\ and quality constraints in the reward function\end{tabular} & DUC2004 & Computationally expensive \\
 & \citet{kwon2023abstractive} & \begin{tabular}[c]{@{}l@{}}Summary length prediction task on the encoder side \\ and encoded this information inserting a \\ length-fusion positional encoding layer\end{tabular} & CNNDM, NYT, WikiHow & \begin{tabular}[c]{@{}l@{}}Performance decreases with increase in summary \\ length variance\end{tabular} \\  
 & \cite{zhang2023macsum} & Hard prompt tuning and soft prefix tuning & CNNDM, QMSum & Low specificity in long generated summaries \\ \midrule
\multirow{6}{*}{Topic} & \citet{krishna-srinivasan-2018-generating} & \begin{tabular}[c]{@{}l@{}}RNN based attention model to generate multiple topic \\ conditioned summaries\end{tabular} & CNNDM & \begin{tabular}[c]{@{}l@{}} News categories provide predefined topics, \\ limiting generalization to other tasks.\end{tabular} \\  
 & \citet{tan-etal-2020-summarizing} & \begin{tabular}[c]{@{}l@{}}Extends topic based summarization to arbitrary topics, \\ integrating external knowledge from ConceptNet and Wikipedia\end{tabular} & \begin{tabular}[c]{@{}l@{}}CNNDM, MA News,\\  All the News\end{tabular} & - \\  
 & \citet{suhara-etal-2020-opiniondigest} & Framework for opinion summarization & HOTEL, Yelp & - \\ 
 & \citet{amplayo-etal-2021-aspect} & \begin{tabular}[c]{@{}l@{}}Multi-Instance Learning and a document preprocessing \\ mechanism\end{tabular} & SPACE, OPOSUM+ & Incapable of handling unseen aspects \\  
 & \citet{mukherjee2020read}  & Iterative sentence extraction algorithm & YELP & Poor performance in absence of attributes \\  
 & \citet{mukherjee-etal-2022-topic} & Topic-aware multimodal summarization system & MSMO & Output quality relies on data size \\ \bottomrule
\end{tabular}%
}
\caption{CTS models descriptions and corresponding limitations.}
\label{tab:models_descritpion}
\end{table*}

\section{Survey papers checklist explanation}
\label{sec:survey_check_list}
\begin{figure}[ht]
    \centering
    \pgfplotsset{every tick label/.append style={font=\scriptsize}}
\pgfplotsset{xlabel/.append style={font=\footnotesize}}
\pgfplotsset{ylabel/.append style={font=\footnotesize}}
\pgfplotstableread{
        Source_of_the_data        Frequency
        Documents                    47
        Reviews                       6
        Dialogues                     5 
        Documents-Dialogues           2  
        Wikipedia                     1
}\datalabel

\begin{tikzpicture}
    \begin{axis}[
        ybar,
        width=0.5\textwidth,
        height=0.35\textwidth,
        ylabel style={inner sep=0ex},
        xtick=data,
        nodes near coords,
        yticklabel style={rotate=75},
        xlabel={Source of the data},
        xticklabels={Documents, Reviews, Dialogues, \begin{tabular}[c]{@{}l@{}} Documents\& \\Dialogues\end{tabular}, Wikipedia} ,
        ymin=0,
        ymax=55,
        xtick style={draw=none},
        bar width=1.5ex,
        xlabel style={inner sep=0ex},
        xtick align=inside,
        ylabel={Frequency},
        nodes near coords style={font=\scriptsize, rotate=90, anchor=east, xshift=13pt}
    ]
    \addplot[] table [y=Frequency, x expr=\coordindex] {\datalabel};
    \end{axis}
\end{tikzpicture}
    \caption{Source of the datasets used for the CTS task. The majority of the data samples are of `document' type.}
    \label{fig:source_data}
\end{figure}
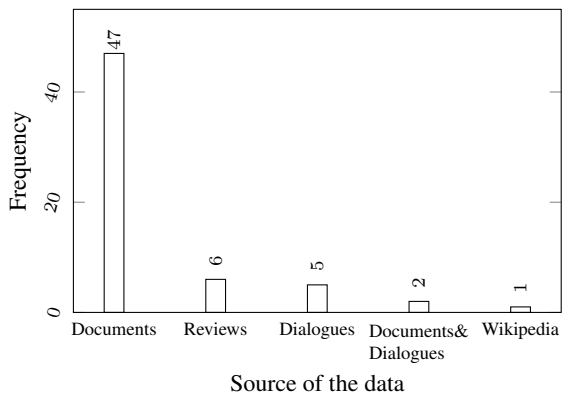

To underscore the comprehensiveness of our survey, as mentioned in Table~\ref{tab:checklist}, we include 23 features for each paper. For easier understanding, we briefly describe each feature in the master table below.
\begin{itemize}
    \setlength{\itemsep}{0pt}
    \item \textit{Paper:} Citation of the paper.
    \item  \textit{Year:} Year of the publication.
    \item \textit{Venue:} Paper publishing conference or journal.
    \item \textit{Controllable attribute:} Controllable attribute(s) concentrated in the paper.
    \item \textit{Controlling more than one aspect:} Whether the paper handles more than one controllable aspect or not?
    \item \textit{Model type:} Type of the model used in the paper such as encoder-decoder, encoder, or decoder architecture.
    \item \textit{Training strategy:} Training approaches employed to perform CTS task.
    \item \textit{Approach:} Type of the training approach employed to perform CTS task.
    \item \textit{Code access:} Whether the code is publicly accessible or not?
    \item \textit{Code link:}  Address of the public repository.
    \item \textit{Dataset:} Dataset utilized in the paper.
    \item \textit{Source:} Source of the dataset used in the paper.
    \item \textit{Nature of the data:} Dataset creation/acquisition strategy.
    \item \textit{Data release:} Public availability of the dataset.
    \item \textit{Domain:} The corresponding domain of the dataset.
    \item \textit{Data link:} Public repository link to the dataset.
    \item \textit{Metric name:} Name of the metric used in the paper.
    \item \textit{Proposed new metric:}  Names of the proposed new automatic evaluation metrics.
    \item \textit{Human evaluation:} Human evaluation performed or not?
    \item \textit{Metric names:} Name of the metrics used to perform human evaluation.
    \item \textit{IAA:} Whether Inter Annotator Agreement assessment performed or not?
    \item \textit{Limitation:} Any limitations of the paper mentioned or not?
    \item \textit{Reproducibility:} Rate the reproducibility of the paper.
\end{itemize}
From the master table, we have represented our observations in Figures~\ref{fig:publication_year}, \ref{fig:type_of_method}, \ref{fig:domain_stats}, \ref{fig:model_type}, \ref{fig:source_data}.  
\label{sec:papers_checklist}
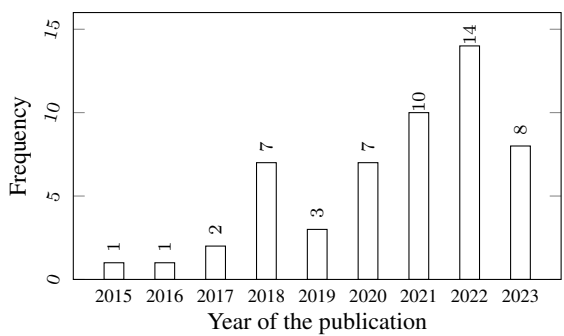
\begin{figure}[ht]
    \centering
    \pgfplotsset{every tick label/.append style={font=\scriptsize}}
\pgfplotsset{xlabel/.append style={font=\footnotesize}}
\pgfplotsset{ylabel/.append style={font=\footnotesize}}
\pgfplotstableread{
        Year    Frequency
        2015      1
        2016      1
        2017      2
        2018      7
        2019      3
        2020      7
        2021      10
        2022      14
        2023      8
}\datalabel

\begin{tikzpicture}
    \begin{axis}[
        ybar,
        width=0.50\textwidth,
        height=0.32\textwidth,
        ylabel style={inner sep=0ex},
        xtick=data,
        nodes near coords,
        yticklabel style={rotate=75},
        xlabel={Year of the publication},
        xticklabels={2015,2016,2017,2018,2019,2020,2021,2022,2023},
        ymin=0,
        ymax=16,
        xtick style={draw=none},
        bar width=1.5ex,
        xlabel style={inner sep=0ex},
        xtick align=inside,
        ylabel={Frequency},
        nodes near coords style={font=\scriptsize,rotate=90, anchor=east,  xshift=12pt}
    ]
    \addplot[] table [y=Frequency, x expr=\coordindex] {\datalabel};
    \end{axis}
\end{tikzpicture}
    \caption{Year-wise papers published in CTS to handle various controllable attributes.}
    \label{fig:publication_year}
\end{figure}
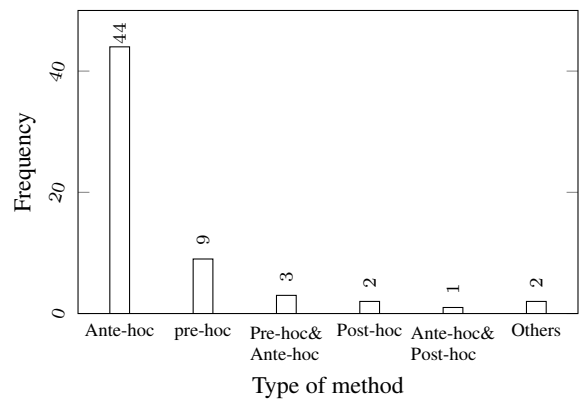
\begin{figure}[ht]
    \centering
    \pgfplotsset{every tick label/.append style={font=\scriptsize}}
\pgfplotsset{xlabel/.append style={font=\footnotesize}}
\pgfplotsset{ylabel/.append style={font=\footnotesize}}
\pgfplotstableread{
        Type_of_method         Frequency
        Ante-hoc               44
        Pre-hoc                9
        Pre-hoc_and_Ante-hoc   3
        Post-hoc               2  
        Ante-hoc_and_Post-hoc  1
        Others                 2 
}\datalabel

\begin{tikzpicture}
    \begin{axis}[
        ybar,
        width=0.51\textwidth,
        height=0.35\textwidth,
        ylabel style={inner sep=0ex},
        xtick=data,
        nodes near coords,
        yticklabel style={rotate=75},
        xlabel={Type of method},
        xticklabels={Ante-hoc, pre-hoc, \begin{tabular}[c]{@{}l@{}} Pre-hoc\& \\Ante-hoc\end{tabular},Post-hoc,\begin{tabular}[c]{@{}l@{}} Ante-hoc\&\\Post-hoc\end{tabular},Others} ,
        ymin=0,
        ymax=50,
        xtick style={draw=none},
        bar width=1.5ex,
        xlabel style={inner sep=0ex},
        xtick align=inside,
        ylabel={Frequency},
        nodes near coords style={font=\scriptsize, rotate=90, anchor=east, xshift=13pt}
    ]
    \addplot[] table [y=Frequency, x expr=\coordindex] {\datalabel};
    \end{axis}
\end{tikzpicture}
    \caption{Various training approaches utilized to perform CTS tasks.}
    \label{fig:type_of_method}
\end{figure}
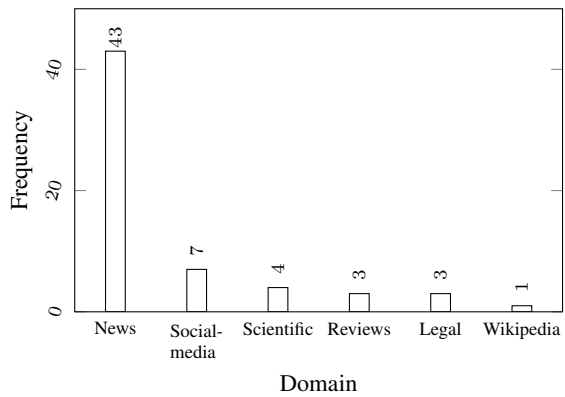
\begin{figure}[ht]
    \centering
    \pgfplotsset{every tick label/.append style={font=\scriptsize}}
\pgfplotsset{xlabel/.append style={font=\footnotesize}}
\pgfplotsset{ylabel/.append style={font=\footnotesize}}
\pgfplotstableread{
        Domain                Frequency
        News                   43
        Social-media            7
        Scientific              4
        Reviews                 3  
        Legal                   3
        Wikipedia               1 
}\datalabel

\begin{tikzpicture}
    \begin{axis}[
        ybar,
        width=0.5\textwidth,
        height=0.35\textwidth,
        ylabel style={inner sep=0ex},
        xtick=data,
        nodes near coords,
        yticklabel style={rotate=75},
        xlabel={Domain},
        xticklabels={News,\begin{tabular}[c]{@{}l@{}}Social-\\media\end{tabular},Scientific,Reviews,Legal,Wikipedia},
        ymin=0,
        ymax=50,
        xtick style={draw=none},
        bar width=1.5ex,
        xlabel style={inner sep=0ex},
        xtick align=inside,
        ylabel={Frequency},
        nodes near coords style={font=\scriptsize, rotate=90, anchor=east, xshift=13}
    ]
    \addplot[] table [y=Frequency, x expr=\coordindex] {\datalabel};
    \end{axis}
\end{tikzpicture}
    \caption{Domains utilized in CTS; most of the existing CTS tasks build on news domain data due to ease in accessibility.}
    \label{fig:domain_stats}
\end{figure}
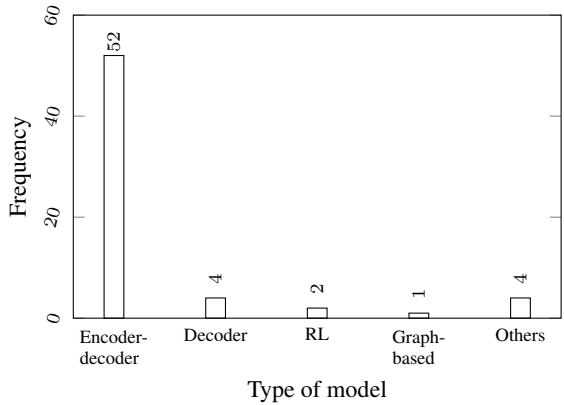
\begin{figure}[htb]
    \centering
    \pgfplotsset{every tick label/.append style={font=\scriptsize}}
\pgfplotsset{xlabel/.append style={font=\footnotesize}}
\pgfplotsset{ylabel/.append style={font=\footnotesize}}
\pgfplotstableread{
        Type_of_model         Frequency
        Encoder-decoder              52
        Decoder                       4
        RL                            2 
        Graph-based                   1  
        Others                        4
}\datalabel

\begin{tikzpicture}
    \begin{axis}[
        ybar,
        width=0.50\textwidth,
        height=0.35\textwidth,
        ylabel style={inner sep=0ex},
        xtick=data,
        nodes near coords,
        yticklabel style={rotate=75},
        xlabel={Type of model},
        xticklabels={\begin{tabular}[c]{@{}l@{}} Encoder- \\decoder\end{tabular}, Decoder,RL,\begin{tabular}[c]{@{}l@{}} Graph-\\based\end{tabular},Others} ,
        ymin=0,
        ymax=60,
        xtick style={draw=none},
        bar width=1.5ex,
        xlabel style={inner sep=0ex},
        xtick align=inside,
        ylabel={Frequency},
        nodes near coords style={font=\scriptsize, rotate=90, anchor=east, xshift=13pt}
    ]
    \addplot[] table [y=Frequency, x expr=\coordindex] {\datalabel};
    \end{axis}
\end{tikzpicture}
    \caption{Type of models used in CTS; the majority of the models fall under standard sequence-to-sequence architecture.}
    \label{fig:model_type}
\end{figure}


\begin{sidewaystable}[htb]
\vspace{50ex}
\centering
\resizebox{\textwidth}{!}{%
\begin{tabular}{clrlllllllllllllllllllll}
\toprule
\multirow{2}{*}{{\ul \textbf{}}} & \multicolumn{1}{c}{\multirow{2}{*}{\textbf{Paper}}} & \multicolumn{2}{c}{\textbf{Publication info}} & \multicolumn{2}{c}{\textbf{Controllable aspect}} & \multicolumn{4}{c}{\textbf{Model details}} & \multicolumn{7}{c}{\textbf{Dataset details}} & \multicolumn{2}{c}{\textbf{Automatic evaluation}} & \multicolumn{3}{c}{\textbf{Human Evaluation}} & \textbf{Limitations} & \textbf{Reproducibility} \\ 
 & \multicolumn{1}{c}{} & \multicolumn{1}{l}{\textbf{Year}} & \textbf{Venue} & \multicolumn{1}{l}{\textbf{\begin{tabular}[c]{@{}l@{}}Controllable \\ attribute\end{tabular}}} & \textbf{\begin{tabular}[c]{@{}l@{}}Controlling \\ more than \\ one aspect\end{tabular}} & \multicolumn{1}{l}{\textbf{Model type*}} & \multicolumn{1}{l}{\textbf{Training strategy}} & \multicolumn{1}{l}{\textbf{Approach}} & \textbf{Code access} & \multicolumn{1}{l}{\textbf{Code link}} & \multicolumn{1}{l}{\textbf{Dataset}} & \multicolumn{1}{l}{\textbf{Source}} & \multicolumn{1}{l}{\textbf{Nature of the data}} & \multicolumn{1}{l}{\textbf{Data release}} & \multicolumn{1}{l}{\textbf{Domain}} & \textbf{Data link} & \multicolumn{1}{l}{\textbf{Metric name}} & \textbf{Proposed New Metric} & \multicolumn{1}{l}{\textbf{Human Evaluation}} & \multicolumn{1}{l}{\textbf{Metrics names}} & \textbf{IAA Yes or No} & \textbf{Mentioned?} & \textbf{High/Medium/Not sure*} \\ 
 \cmidrule(lr){3-4}\cmidrule(lr){5-6}\cmidrule(lr){7-11}\cmidrule(lr){12-17}\cmidrule(lr){18-19}\cmidrule(lr){20-22}\cmidrule(lr){23-23}\cmidrule(lr){24-24} \\
1 & \citet{chawla-etal-2019-generating} & \multicolumn{1}{r}{2019} & CoNLL & \multicolumn{1}{l}{Style} & no & \multicolumn{1}{l}{encoder-decoder} & \multicolumn{1}{l}{ante-hoc} & \multicolumn{1}{l}{supervised} & no & \multicolumn{1}{l}{-} & \multicolumn{1}{l}{\begin{tabular}[c]{@{}l@{}}CNNDM, \\ Webis-TLDR-17 corpus\end{tabular}} & \multicolumn{1}{l}{document} & \multicolumn{1}{l}{generic} & \multicolumn{1}{l}{no} & \multicolumn{1}{l}{news, social media} & no & \multicolumn{1}{l}{ROUGE} & no & \multicolumn{1}{l}{no} & \multicolumn{1}{l}{no} & no & no & not sure \\ 
2 & \citet{goyal-etal-2022-hydrasum} & \multicolumn{1}{r}{2022} & EMNLP & \multicolumn{1}{l}{Style} & yes & \multicolumn{1}{l}{encoder-decoder} & \multicolumn{1}{l}{ante-hoc} & \multicolumn{1}{l}{supervised} & yes & \multicolumn{1}{l}{{\ul \href{https://github.com/salesforce/hydra-sum}{Link}}} & \multicolumn{1}{l}{\begin{tabular}[c]{@{}l@{}}CNNDM,\\  XSum, NEWSROOM\end{tabular}} & \multicolumn{1}{l}{document} & \multicolumn{1}{l}{generic} & \multicolumn{1}{l}{no} & \multicolumn{1}{l}{news} & no & \multicolumn{1}{l}{ROUGE} & \begin{tabular}[c]{@{}l@{}}abstractiveness, \\ degree of specificity, \\ length, readability\end{tabular} & \multicolumn{1}{l}{yes} & \multicolumn{1}{l}{\begin{tabular}[c]{@{}l@{}}relevance, coherence, \\ grammaticality, \\ factuality\end{tabular}} & no & yes & high \\ 
3 & \citet{song-etal-2021-new} & \multicolumn{1}{r}{2021} & NAACL & \multicolumn{1}{l}{Length} & no & \multicolumn{1}{l}{decoder} & \multicolumn{1}{l}{ante-hoc} & \multicolumn{1}{l}{supervised} & yes & \multicolumn{1}{l}{{\ul \href{https://github.com/ucfnlp/varying-length-summ}{Link}}} & \multicolumn{1}{l}{Gigaword, NEWSROOM} & \multicolumn{1}{l}{document} & \multicolumn{1}{l}{generic} & \multicolumn{1}{l}{yes} & \multicolumn{1}{l}{news} & {\ul \href{https://github.com/ucfnlp/varying-length-summ/tree/main/data}{Link}} & \multicolumn{1}{l}{ROUGE} & no & \multicolumn{1}{l}{yes} & \multicolumn{1}{l}{yes-no type questions} & no & no & high \\ 
4 & \citet{song2020controlling} & \multicolumn{1}{r}{2020} & AAAI & \multicolumn{1}{l}{Abstractivity} & no & \multicolumn{1}{l}{decoder} & \multicolumn{1}{l}{ante-hoc} & \multicolumn{1}{l}{supervised} & yes & \multicolumn{1}{l}{{\ul \href{https://github.com/ucfnlp/control-over-copying}{Link}}} & \multicolumn{1}{l}{Gigaword, NEWSROOM} & \multicolumn{1}{l}{document} & \multicolumn{1}{l}{generic} & \multicolumn{1}{l}{no} & \multicolumn{1}{l}{news} & no & \multicolumn{1}{l}{ROUGE, BERTScore} & no & \multicolumn{1}{l}{yes} & \multicolumn{1}{l}{\begin{tabular}[c]{@{}l@{}}informativeness, \\ grammaticality, \\ truthfulness\end{tabular}} & no & no & high \\ 
5 & \citet{fan2018controllable} & \multicolumn{1}{r}{2018} & ACL-NMT(W) & \multicolumn{1}{l}{\begin{tabular}[c]{@{}l@{}}Length, Style\\ Entity\end{tabular}} & yes & \multicolumn{1}{l}{encoder-decoder} & \multicolumn{1}{l}{pre-hoc} & \multicolumn{1}{l}{supervised} & no & \multicolumn{1}{l}{-} & \multicolumn{1}{l}{CNNDM} & \multicolumn{1}{l}{document} & \multicolumn{1}{l}{generic} & \multicolumn{1}{l}{no} & \multicolumn{1}{l}{news} & no & \multicolumn{1}{l}{ROUGE} & no & \multicolumn{1}{l}{yes} & \multicolumn{1}{l}{\begin{tabular}[c]{@{}l@{}}rater asked to preferred summary, \\ binary evaluation\end{tabular}} & no & no & medium \\ 
6 & \citet{liu-etal-2022-length} & \multicolumn{1}{r}{2022} & ACL & \multicolumn{1}{l}{Length} & no & \multicolumn{1}{l}{encoder-decoder} & \multicolumn{1}{l}{ante-hoc} & \multicolumn{1}{l}{supervised} & yes & \multicolumn{1}{l}{{\ul \href{https://github.com/yizhuliu/lengthcontrol}{Link}}} & \multicolumn{1}{l}{CNNDM, XSum} & \multicolumn{1}{l}{document} & \multicolumn{1}{l}{generic} & \multicolumn{1}{l}{no} & \multicolumn{1}{l}{news} & no & \multicolumn{1}{l}{ROUGE} & no & \multicolumn{1}{l}{yes} & \multicolumn{1}{l}{\begin{tabular}[c]{@{}l@{}}grammatically correct, \\ informativeness, overall\end{tabular}} & no & no & high \\ 
7 & \citet{zhang2023macsum} & \multicolumn{1}{r}{2023} & TACL & \multicolumn{1}{l}{\begin{tabular}[c]{@{}l@{}}Length, Topic \\ Coverage\end{tabular}} & yes & \multicolumn{1}{l}{encoder-decoder} & \multicolumn{1}{l}{pre-hoc} & \multicolumn{1}{l}{supervised} & yes & \multicolumn{1}{l}{{\ul \href{https://github.com/yizhuliu/lengthcontrol}{Link}}} & \multicolumn{1}{l}{CNNDM, QMSum} & \multicolumn{1}{l}{document, dialogues} & \multicolumn{1}{l}{human annotated} & \multicolumn{1}{l}{yes} & \multicolumn{1}{l}{news, social media} & {\ul \href{https://huggingface.co/datasets/chatc/MACSum}{Link}} & \multicolumn{1}{l}{\begin{tabular}[c]{@{}l@{}}Control Error Rate, \\ Control Correlation\end{tabular}} & \begin{tabular}[c]{@{}l@{}}control error rate,\\ control correlation\end{tabular} & \multicolumn{1}{l}{yes} & \multicolumn{1}{l}{yes or no questions} & yes & no & high \\ 
8 & \citet{takase-okazaki-2019-positional} & \multicolumn{1}{r}{2019} & NAACL & \multicolumn{1}{l}{Length} & no & \multicolumn{1}{l}{encoder-decoder} & \multicolumn{1}{l}{ante-hoc} & \multicolumn{1}{l}{supervised} & yes & \multicolumn{1}{l}{{\ul \href{https://github.com/takase/control-length}{Link}}} & \multicolumn{1}{l}{JAMUS} & \multicolumn{1}{l}{document} & \multicolumn{1}{l}{generic} & \multicolumn{1}{l}{no} & \multicolumn{1}{l}{news} & no & \multicolumn{1}{l}{ROUGE} & no & \multicolumn{1}{l}{no} & \multicolumn{1}{l}{no} & no & no & medium \\ 
9 & \citet{shen-etal-2022-sentbs} & \multicolumn{1}{r}{2022} & EMNLP & \multicolumn{1}{l}{Structure} & no & \multicolumn{1}{l}{encoder-decoder} & \multicolumn{1}{l}{post-hoc} & \multicolumn{1}{l}{supervised} & yes & \multicolumn{1}{l}{{\ul \href{https://github.com/Shen-Chenhui/SentBS}{Link}}} & \multicolumn{1}{l}{Meta Review Dataset (MReD)} & \multicolumn{1}{l}{reviews} & \multicolumn{1}{l}{generic} & \multicolumn{1}{l}{no} & \multicolumn{1}{l}{scientific} & no & \multicolumn{1}{l}{ROUGE,BERTSCORE} & no & \multicolumn{1}{l}{yes} & \multicolumn{1}{l}{\begin{tabular}[c]{@{}l@{}}fluency, content relevance, \\ structure similarity, \\ decision correctness\end{tabular}} & no & yes & high \\ 
10 & \citet{goldsack-etal-2023-biolaysumm} & \multicolumn{1}{r}{2023} & ACL-BIoNLP(W) & \multicolumn{1}{l}{Style} & no & \multicolumn{1}{l}{-} & \multicolumn{1}{l}{-} & \multicolumn{1}{l}{-} & - & \multicolumn{1}{l}{-} & \multicolumn{1}{l}{PLOS and eLife} & \multicolumn{1}{l}{-} & \multicolumn{1}{l}{-} & \multicolumn{1}{l}{-} & \multicolumn{1}{l}{-} & - & \multicolumn{1}{l}{\begin{tabular}[c]{@{}l@{}}Relevance, Readability,\\  Factuality\end{tabular}} & - & \multicolumn{1}{l}{-} & \multicolumn{1}{l}{-} & no & - & - \\ 
11 & \citet{cao-wang-2021-inference} & \multicolumn{1}{r}{2021} & NAACL & \multicolumn{1}{l}{Style} & no & \multicolumn{1}{l}{encoder-decoder} & \multicolumn{1}{l}{ante-hoc} & \multicolumn{1}{l}{supervised} & yes & \multicolumn{1}{l}{{\ul \href{https://github.com/ShuyangCao/inference\_style\_control}{Link}}} & \multicolumn{1}{l}{CNNDM} & \multicolumn{1}{l}{document} & \multicolumn{1}{l}{generic} & \multicolumn{1}{l}{no} & \multicolumn{1}{l}{news} & no & \multicolumn{1}{l}{\begin{tabular}[c]{@{}l@{}}Readability, Perplexity, \\ BERTScore\end{tabular}} & \begin{tabular}[c]{@{}l@{}}readability, \\ perplexity\end{tabular} & \multicolumn{1}{l}{yes} & \multicolumn{1}{l}{\begin{tabular}[c]{@{}l@{}}informativeness, \\ fluency, relevance\end{tabular}} & no & no & high \\ 
12 & \citet{narayan-etal-2021-planning} & \multicolumn{1}{r}{2021} & TACL & \multicolumn{1}{l}{Structure} & no & \multicolumn{1}{l}{encoder-decoder} & \multicolumn{1}{l}{ante-hoc} & \multicolumn{1}{l}{supervised} & no & \multicolumn{1}{l}{-} & \multicolumn{1}{l}{\begin{tabular}[c]{@{}l@{}}CNNDM, XSum, \\ SAMSum, and BillSum\end{tabular}} & \multicolumn{1}{l}{document, dialogues} & \multicolumn{1}{l}{generic} & \multicolumn{1}{l}{no} & \multicolumn{1}{l}{news, legal} & no & \multicolumn{1}{l}{ROUGE} & \begin{tabular}[c]{@{}l@{}}entity planning, \\ specificity\end{tabular} & \multicolumn{1}{l}{yes} & \multicolumn{1}{l}{\begin{tabular}[c]{@{}l@{}}relevant, accurate, \\ concise, fluent\end{tabular}} & no & no & medium \\ 
13 & \citet{amplayo-etal-2021-aspect} & \multicolumn{1}{r}{2021} & EMNLP & \multicolumn{1}{l}{Topic} & no & \multicolumn{1}{l}{encoder-decoder} & \multicolumn{1}{l}{ante-hoc} & \multicolumn{1}{l}{supervised} & yes & \multicolumn{1}{l}{{\ul \href{https://github.com/rktamplayo/AceSum}{Link}}} & \multicolumn{1}{l}{SPACE, OPOSUM+} & \multicolumn{1}{l}{reviews} & \multicolumn{1}{l}{generic} & \multicolumn{1}{l}{no} & \multicolumn{1}{l}{social media} & no & \multicolumn{1}{l}{ROUGE} & no & \multicolumn{1}{l}{yes} & \multicolumn{1}{l}{\begin{tabular}[c]{@{}l@{}}informativeness,coherence,\\  conciseness, fluency\end{tabular}} & no & no & high \\ 
14 & \citet{deutsch2023incorporating} & \multicolumn{1}{r}{2023} & EACL & \multicolumn{1}{l}{Saliency} & no & \multicolumn{1}{l}{encoder-decoder} & \multicolumn{1}{l}{ante-hoc} & \multicolumn{1}{l}{supervised} & no & \multicolumn{1}{l}{-} & \multicolumn{1}{l}{CNNDM, XSum, NYTimes} & \multicolumn{1}{l}{document} & \multicolumn{1}{l}{generic} & \multicolumn{1}{l}{no} & \multicolumn{1}{l}{news} & no & \multicolumn{1}{l}{\begin{tabular}[c]{@{}l@{}}ROUGE, BERTScore, \\ QAEval\end{tabular}} & no & \multicolumn{1}{l}{yes} & \multicolumn{1}{l}{\begin{tabular}[c]{@{}l@{}}selection of best summary \\ (binary evaluation)\end{tabular}} & no & no & not sure \\ 
15 & \citet{wu-etal-2021-controllable} & \multicolumn{1}{r}{2021} & ALC-IJCNLP & \multicolumn{1}{l}{Coverage} & no & \multicolumn{1}{l}{encoder-decoder} & \multicolumn{1}{l}{ante-hoc} & \multicolumn{1}{l}{supervised} & yes & \multicolumn{1}{l}{{\ul \href{https://github.com/salesforce/ConvSumm}{Link}}} & \multicolumn{1}{l}{SAMSum} & \multicolumn{1}{l}{dialogues} & \multicolumn{1}{l}{generic} & \multicolumn{1}{l}{no} & \multicolumn{1}{l}{social media} & no & \multicolumn{1}{l}{ROUGE} & no & \multicolumn{1}{l}{yes} & \multicolumn{1}{l}{\begin{tabular}[c]{@{}l@{}}factual consistency,\\  informativeness\end{tabular}} & no & no & high \\ 
16 & \citet{hsu-tan-2021-decision}  & \multicolumn{1}{r}{2021} & EMNLP & \multicolumn{1}{l}{Topic} & no & \multicolumn{1}{l}{encoder-decoder} & \multicolumn{1}{l}{ante-hoc} & \multicolumn{1}{l}{supervised} & yes & \multicolumn{1}{l}{{\ul \href{https://github.com/ChicagoHAI/decsum}{Link}}} & \multicolumn{1}{l}{\begin{tabular}[c]{@{}l@{}}Yelp's businesses, reviews, \\ and user data\end{tabular}} & \multicolumn{1}{l}{reviews} & \multicolumn{1}{l}{generic} & \multicolumn{1}{l}{no} & \multicolumn{1}{l}{restaurant reviews} & {\ul \href{https://github.com/ChicagoHAI/decsum}{Link}} & \multicolumn{1}{l}{MSE} & no & \multicolumn{1}{l}{yes} & \multicolumn{1}{l}{-} & no & no & high \\ 
17 & \citet{hyun-etal-2022-generating} & \multicolumn{1}{r}{2022} & EMNLP & \multicolumn{1}{l}{Length} & yes & \multicolumn{1}{l}{RL} & \multicolumn{1}{l}{ante-hoc} & \multicolumn{1}{l}{unsupervised} & yes & \multicolumn{1}{l}{{\ul \href{https://github.com/dmhyun/MSRP}{Link}}} & \multicolumn{1}{l}{DUC2004} & \multicolumn{1}{l}{document} & \multicolumn{1}{l}{generic} & \multicolumn{1}{l}{no} & \multicolumn{1}{l}{news} & no & \multicolumn{1}{l}{ROUGE} & no & \multicolumn{1}{l}{yes} & \multicolumn{1}{l}{fluency, fidelity} & yes & yes & high \\ 
18 & \citet{Chan2021ControllableSW} & \multicolumn{1}{r}{2021} & TACL & \multicolumn{1}{l}{\begin{tabular}[c]{@{}l@{}}Length, Entity, \\ Abstractivity\end{tabular}} & yes & \multicolumn{1}{l}{RL} & \multicolumn{1}{l}{ante-hoc} & \multicolumn{1}{l}{supervised} & yes & \multicolumn{1}{l}{{\ul \href{https://github.com/kenchan0226/control-sum-cmdp}{Link}}} & \multicolumn{1}{l}{CNNDM, NEWSROOM, DUC-2002} & \multicolumn{1}{l}{document} & \multicolumn{1}{l}{generic} & \multicolumn{1}{l}{no} & \multicolumn{1}{l}{news} & no & \multicolumn{1}{l}{\begin{tabular}[c]{@{}l@{}}ROUGE, BERTSCORE, \\ MoveScore, QA-F\end{tabular}} & yes & \multicolumn{1}{l}{yes} & \multicolumn{1}{l}{\begin{tabular}[c]{@{}l@{}}fluency, entity relevance,\\ faithfulness\end{tabular}} & yes & no & high \\ 
19 & \citet{krishna-srinivasan-2018-generating} & \multicolumn{1}{r}{2018} & NAACL-HLT & \multicolumn{1}{l}{Topic} & no & \multicolumn{1}{l}{encoder-decoder} & \multicolumn{1}{l}{pre-hoc} & \multicolumn{1}{l}{supervised} & no & \multicolumn{1}{l}{-} & \multicolumn{1}{l}{CNNDM} & \multicolumn{1}{l}{document} & \multicolumn{1}{l}{generic} & \multicolumn{1}{l}{no} & \multicolumn{1}{l}{news} & no & \multicolumn{1}{l}{ROUGE} & no & \multicolumn{1}{l}{yes} & \multicolumn{1}{l}{topic relevance} & no & no & not sure \\ 
20 & \citet{krishna2018vocabulary} & \multicolumn{1}{r}{2018} & ACL & \multicolumn{1}{l}{Topic} & no & \multicolumn{1}{l}{encoder-decoder} & \multicolumn{1}{l}{ante-hoc} & \multicolumn{1}{l}{supervised} & no & \multicolumn{1}{l}{-} & \multicolumn{1}{l}{CNNDM} & \multicolumn{1}{l}{document} & \multicolumn{1}{l}{generic} & \multicolumn{1}{l}{no} & \multicolumn{1}{l}{news} & no & \multicolumn{1}{l}{ROUGE} & no & \multicolumn{1}{l}{yes} & \multicolumn{1}{l}{quality} & yes & no & not sure \\ 
21 & \citet{rush2015neural} & \multicolumn{1}{r}{2015} & EMNLP & \multicolumn{1}{l}{Length} & no & \multicolumn{1}{l}{encoder-decoder} & \multicolumn{1}{l}{ante-hoc} & \multicolumn{1}{l}{supervised} & no & \multicolumn{1}{l}{-} & \multicolumn{1}{l}{NYT, DUC2004} & \multicolumn{1}{l}{document} & \multicolumn{1}{l}{generic} & \multicolumn{1}{l}{no} & \multicolumn{1}{l}{news} & no & \multicolumn{1}{l}{ROUGE} & no & \multicolumn{1}{l}{no} & \multicolumn{1}{l}{no} & no & no & high \\ 
22 & \citet{fevry-phang-2018-unsupervised} & \multicolumn{1}{r}{2018} & CoNLL & \multicolumn{1}{l}{Length} & no & \multicolumn{1}{l}{encoder-decoder} & \multicolumn{1}{l}{ante-hoc} & \multicolumn{1}{l}{unsupervised} & yes & \multicolumn{1}{l}{{\ul \href{https://github.com/zphang/usc\_dae}{Link}}} & \multicolumn{1}{l}{Gigaword} & \multicolumn{1}{l}{document} & \multicolumn{1}{l}{generic} & \multicolumn{1}{l}{no} & \multicolumn{1}{l}{news} & no & \multicolumn{1}{l}{ROUGE} & no & \multicolumn{1}{l}{yes} & \multicolumn{1}{l}{\begin{tabular}[c]{@{}l@{}}grammaticality, \\ preserving original information\end{tabular}} & no & no & high \\ 
23 & \citet{li-etal-2018-guiding} & \multicolumn{1}{r}{2018} & NAACL-HLT & \multicolumn{1}{l}{Saliency} & no & \multicolumn{1}{l}{encoder-decoder} & \multicolumn{1}{l}{ante-hoc} & \multicolumn{1}{l}{supervised} & no & \multicolumn{1}{l}{-} & \multicolumn{1}{l}{CNNDM} & \multicolumn{1}{l}{document} & \multicolumn{1}{l}{generic} & \multicolumn{1}{l}{no} & \multicolumn{1}{l}{news} & no & \multicolumn{1}{l}{ROUGE} & no & \multicolumn{1}{l}{no} & \multicolumn{1}{l}{no} & no & no & not sure \\ 
24 & \citet{jin2020semsum} & \multicolumn{1}{r}{2020} & AAAI & \multicolumn{1}{l}{Coverage} & no & \multicolumn{1}{l}{encoder-decoder} & \multicolumn{1}{l}{ante-hoc} & \multicolumn{1}{l}{supervised} & no & \multicolumn{1}{l}{-} & \multicolumn{1}{l}{\begin{tabular}[c]{@{}l@{}}Gigaword, DUC2004,\\ MSR dataset\end{tabular}} & \multicolumn{1}{l}{document} & \multicolumn{1}{l}{generic} & \multicolumn{1}{l}{no} & \multicolumn{1}{l}{news} & {\ul \href{https://github.com/harvardnlp/sent-summary}{Link}} & \multicolumn{1}{l}{ROUGE} & \begin{tabular}[c]{@{}l@{}}Word Mover's Distance, \\ Bert Score\end{tabular} & \multicolumn{1}{l}{yes} & \multicolumn{1}{l}{\begin{tabular}[c]{@{}l@{}}faithfulness, informativeness, \\ fluency\end{tabular}} & no & no & not sure \\ 
25 & \citet{liu2022character} & \multicolumn{1}{r}{2022} & NeurIPS & \multicolumn{1}{l}{Length} & no & \multicolumn{1}{l}{\begin{tabular}[c]{@{}l@{}}Non Auto Regressive Model\end{tabular}} & \multicolumn{1}{l}{ante-hoc} & \multicolumn{1}{l}{supervised} & yes & \multicolumn{1}{l}{{\ul \href{https://github.com/MANGA-UOFA/NACC}{Link}}} & \multicolumn{1}{l}{Gigaword, DUC2004} & \multicolumn{1}{l}{document} & \multicolumn{1}{l}{generic} & \multicolumn{1}{l}{no} & \multicolumn{1}{l}{news} & no & \multicolumn{1}{l}{ROUGE} & no & \multicolumn{1}{l}{yes} & \multicolumn{1}{l}{\begin{tabular}[c]{@{}l@{}}overall quality, \\ completeness/fluency\end{tabular}} & no & yes & medium \\ 
26 & \citet{kwon2023abstractive} & \multicolumn{1}{r}{2023} & EACL & \multicolumn{1}{l}{Length} & no & \multicolumn{1}{l}{encoder-decoder} & \multicolumn{1}{l}{ante-hoc} & \multicolumn{1}{l}{supervised} & no & \multicolumn{1}{l}{-} & \multicolumn{1}{l}{CNNDM, NYT, WikiHow} & \multicolumn{1}{l}{document} & \multicolumn{1}{l}{generic} & \multicolumn{1}{l}{no} & \multicolumn{1}{l}{news} & no & \multicolumn{1}{l}{ROUGE, Length Variance} & no & \multicolumn{1}{l}{yes} & \multicolumn{1}{l}{\begin{tabular}[c]{@{}l@{}}informativeness, \\ conciseness\end{tabular}} & no & yes & not sure \\ 
27 & \citet{pagnoni-etal-2023-socratic} & \multicolumn{1}{r}{2023} & ACL & \multicolumn{1}{l}{Entity, Saliency} & yes & \multicolumn{1}{l}{encoder-decoder} & \multicolumn{1}{l}{ante-hoc} & \multicolumn{1}{l}{supervised} & yes & \multicolumn{1}{l}{{\ul \href{https://github.com/salesforce/socratic-pretraining}{Link}}} & \multicolumn{1}{l}{QMSum and SQuALITY} & \multicolumn{1}{l}{document} & \multicolumn{1}{l}{generic} & \multicolumn{1}{l}{no} & \multicolumn{1}{l}{news} & no & \multicolumn{1}{l}{\begin{tabular}[c]{@{}l@{}}ROUGE, BERTSCORE, \\ levenshtein distance\end{tabular}} & no & \multicolumn{1}{l}{yes} & \multicolumn{1}{l}{unambiguity, conciseness} & no & yes & high \\ 
28 & \citet{hofmann-coyle-etal-2022-extractive} & \multicolumn{1}{r}{2022} & AACL & \multicolumn{1}{l}{Entity} & no & \multicolumn{1}{l}{encoder-decoder} & \multicolumn{1}{l}{ante-hoc} & \multicolumn{1}{l}{supervised} & no & \multicolumn{1}{l}{-} & \multicolumn{1}{l}{EntSum} & \multicolumn{1}{l}{document} & \multicolumn{1}{l}{generic} & \multicolumn{1}{l}{no} & \multicolumn{1}{l}{news} & no & \multicolumn{1}{l}{F1} & no & \multicolumn{1}{l}{no} & \multicolumn{1}{l}{no} & no & no & not sure \\ 
29 &  \citet{lin-etal-2022-roles} & \multicolumn{1}{r}{2022} & ACL & \multicolumn{1}{l}{Role} & no & \multicolumn{1}{l}{encoder-decoder} & \multicolumn{1}{l}{ante-hoc} & \multicolumn{1}{l}{supervised} & yes & \multicolumn{1}{l}{{\ul \href{https://github.com/xiaolinAndy/RODS}{Link}}} & \multicolumn{1}{l}{CSDS, MC} & \multicolumn{1}{l}{dialogues} & \multicolumn{1}{l}{generic} & \multicolumn{1}{l}{no} & \multicolumn{1}{l}{\begin{tabular}[c]{@{}l@{}}customer service data, \\ medical inquiries\end{tabular}} & {\ul \href{https://github.com/xiaolinAndy/RODS/tree/main/data/MC}{Link}} & \multicolumn{1}{l}{\begin{tabular}[c]{@{}l@{}}ROUGE, BertScore, \\ BLEU, MoverScore\end{tabular}} & no & \multicolumn{1}{l}{yes} & \multicolumn{1}{l}{\begin{tabular}[c]{@{}l@{}}informativeness, \\ non-redundancy, fluency\end{tabular}} & yes & no & high \\ 
30 &\citet{zhong-etal-2022-unsupervised}  & \multicolumn{1}{r}{2022} & EMNLP & \multicolumn{1}{l}{Coverage} & yes & \multicolumn{1}{l}{encoder-decoder} & \multicolumn{1}{l}{ante-hoc} & \multicolumn{1}{l}{unsupervised} & no & \multicolumn{1}{l}{-} & \multicolumn{1}{l}{\begin{tabular}[c]{@{}l@{}}GranuDUC, MultiNews, \\ DUC2004, Arxiv\end{tabular}} & \multicolumn{1}{l}{document} & \multicolumn{1}{l}{generic} & \multicolumn{1}{l}{yes} & \multicolumn{1}{l}{news, scientific papers} & {\ul \href{https://github.com/maszhongming/GranuDUC}{Link}} & \multicolumn{1}{l}{ROUGE} & BERTScore & \multicolumn{1}{l}{yes} & \multicolumn{1}{l}{\begin{tabular}[c]{@{}l@{}}fluency, \\ relevance, faithfulness\end{tabular}} & no & yes & medium \\ 
31 & \citet{liang-etal-2022-towards-modeling} & \multicolumn{1}{r}{2022} & AACL & \multicolumn{1}{l}{Role} & yes & \multicolumn{1}{l}{encoder-decoder} & \multicolumn{1}{l}{ante-hoc} & \multicolumn{1}{l}{supervised} & no & \multicolumn{1}{l}{-} & \multicolumn{1}{l}{CSDS, MC} & \multicolumn{1}{l}{dialogues} & \multicolumn{1}{l}{generic} & \multicolumn{1}{l}{no} & \multicolumn{1}{l}{\begin{tabular}[c]{@{}l@{}}customer service data, \\ medical inquiries\end{tabular}} & {\ul \href{https://github.com/xiaolinAndy/RODS/tree/main/data/MC}{Link}} & \multicolumn{1}{l}{\begin{tabular}[c]{@{}l@{}}ROUGE, BLEU, \\ BERTScore, MoverScore\end{tabular}} & no & \multicolumn{1}{l}{no} & \multicolumn{1}{l}{no} & no & no & not sure \\ 
32 & \citet{liu-chen-2021-controllable} & \multicolumn{1}{r}{2021} & EMNLP & \multicolumn{1}{l}{Entity} & no & \multicolumn{1}{l}{encoder-decoder} & \multicolumn{1}{l}{ante-hoc} & \multicolumn{1}{l}{supervised} & no & \multicolumn{1}{l}{{\ul \href{https://github.com/seq-to-mind/planning\_dial\_summ}{Link}}} & \multicolumn{1}{l}{SAMSum} & \multicolumn{1}{l}{dialogues} & \multicolumn{1}{l}{generic} & \multicolumn{1}{l}{no} & \multicolumn{1}{l}{social media} & no & \multicolumn{1}{l}{ROUGE} & no & \multicolumn{1}{l}{yes} & \multicolumn{1}{l}{\begin{tabular}[c]{@{}l@{}}factual consistency,\\ informativeness\end{tabular}} & no & no & medium \\ 
33 & \citet{zheng-etal-2020-controllable} & \multicolumn{1}{r}{2020} & COLING & \multicolumn{1}{l}{Entity} & no & \multicolumn{1}{l}{encoder-decoder} & \multicolumn{1}{l}{ante-hoc} & \multicolumn{1}{l}{supervised} & yes & \multicolumn{1}{l}{{\ul \href{https://github.com/thecharm/Abs-LRModel/tree/main}{Link}}} & \multicolumn{1}{l}{Gigaword, DUC2004} & \multicolumn{1}{l}{document} & \multicolumn{1}{l}{generic} & \multicolumn{1}{l}{no} & \multicolumn{1}{l}{news} & no & \multicolumn{1}{l}{BERT-REO, ROUGE} & BERT-REO & \multicolumn{1}{l}{yes} & \multicolumn{1}{l}{\begin{tabular}[c]{@{}l@{}}informativeness,\\ grammaticality,\\ coherence\end{tabular}} & no & no & high \\ 
34 & \citet{huang-etal-2023-swing} & \multicolumn{1}{r}{2023} & EACL & \multicolumn{1}{l}{Coverage} & no & \multicolumn{1}{l}{encoder-decoder} & \multicolumn{1}{l}{ante-hoc} & \multicolumn{1}{l}{supervised} & yes & \multicolumn{1}{l}{{\ul \href{https://github.com/amazon-science/AWS-SWING}{Link}}} & \multicolumn{1}{l}{DIALOG-SUM, SAMSUM} & \multicolumn{1}{l}{dialogues} & \multicolumn{1}{l}{generic} & \multicolumn{1}{l}{no} & \multicolumn{1}{l}{social media} & no & \multicolumn{1}{l}{\begin{tabular}[c]{@{}l@{}}BARTScore, ROUGE,\\  FactCC\end{tabular}} & no & \multicolumn{1}{l}{yes} & \multicolumn{1}{l}{\begin{tabular}[c]{@{}l@{}}recall, precision,\\  faithfulness\end{tabular}} & yes & yes & high \\ 
35 & \citet{mukherjee-etal-2022-topic} & \multicolumn{1}{r}{2022} & AACL & \multicolumn{1}{l}{Topic} & no & \multicolumn{1}{l}{encoder-decoder} & \multicolumn{1}{l}{ante-hoc} & \multicolumn{1}{l}{supervised} & yes & \multicolumn{1}{l}{{\ul \href{https://github.com/mailsourajit25/Topic-Aware-Multimodal-Summarization}{Link}}} & \multicolumn{1}{l}{MSMO} & \multicolumn{1}{l}{document} & \multicolumn{1}{l}{semi-automatic} & \multicolumn{1}{l}{yes} & \multicolumn{1}{l}{news} & \href{https://drive.google.com/file/d/1yzEE\_n5q2VaqMrI1q9BrTS9ebrF-nQko/view?usp=sharing}{Link} & \multicolumn{1}{l}{ROUGE} & no & \multicolumn{1}{l}{yes} & \multicolumn{1}{l}{\begin{tabular}[c]{@{}l@{}}Coverage, Grammar, \\ Topic-Aware-Text\end{tabular}} & no & yes & high \\ 
36 & \citet{jin-etal-2020-hooks} & \multicolumn{1}{r}{2020} & ACL & \multicolumn{1}{l}{Style} & yes & \multicolumn{1}{l}{encoder-decoder} & \multicolumn{1}{l}{ante-hoc} & \multicolumn{1}{l}{supervised} & yes & \multicolumn{1}{l}{{\ul \href{https://github.com/jind11/TitleStylist}{Link}}} & \multicolumn{1}{l}{NYT, CNN} & \multicolumn{1}{l}{document} & \multicolumn{1}{l}{generic} & \multicolumn{1}{l}{no} & \multicolumn{1}{l}{news} & no & \multicolumn{1}{l}{\begin{tabular}[c]{@{}l@{}}BLEU, ROUGE,\\ METEOR, CIDEr\end{tabular}} & no & \multicolumn{1}{l}{yes} & \multicolumn{1}{l}{\begin{tabular}[c]{@{}l@{}}relevance, attractiveness, \\ language fluency,\\  style strength\end{tabular}} & no & no & high \\ 
37 & \citet{see-etal-2017-get} & \multicolumn{1}{r}{2017} & ACL & \multicolumn{1}{l}{Abstractivity, Coverage} & yes & \multicolumn{1}{l}{encoder-decoder} & \multicolumn{1}{l}{ante-hoc} & \multicolumn{1}{l}{supervised} & yes & \multicolumn{1}{l}{{\ul \href{https://github.com/abisee/pointer-generator}{Link}}} & \multicolumn{1}{l}{CNNDM} & \multicolumn{1}{l}{document} & \multicolumn{1}{l}{generic} & \multicolumn{1}{l}{no} & \multicolumn{1}{l}{news} & no & \multicolumn{1}{l}{ROUGE} & no & \multicolumn{1}{l}{no} & \multicolumn{1}{l}{no} & no & no & high \\ 
38 & \citet{kryscinski-etal-2018-improving} & \multicolumn{1}{r}{2018} & EMNLP & \multicolumn{1}{l}{Abstractivity} & no & \multicolumn{1}{l}{encoder-decoder} & \multicolumn{1}{l}{ante-hoc} & \multicolumn{1}{l}{supervised} & no & \multicolumn{1}{l}{-} & \multicolumn{1}{l}{CNNDM} & \multicolumn{1}{l}{document} & \multicolumn{1}{l}{generic} & \multicolumn{1}{l}{no} & \multicolumn{1}{l}{news} & no & \multicolumn{1}{l}{ROUGE} & no & \multicolumn{1}{l}{yes} & \multicolumn{1}{l}{readibility, relevance} & no & no & not sure \\ 
39 & \citet{bahrainian2021cats} & \multicolumn{1}{r}{2021} & ACM & \multicolumn{1}{l}{Topic} & no & \multicolumn{1}{l}{encoder-decoder} & \multicolumn{1}{l}{ante-hoc} & \multicolumn{1}{l}{supervised} & yes & \multicolumn{1}{l}{{\ul \href{https://github.com/ali-bahrainian/CATS}{Link}}} & \multicolumn{1}{l}{CNNDM, AMI , ICSI, ADSE} & \multicolumn{1}{l}{document} & \multicolumn{1}{l}{generic} & \multicolumn{1}{l}{no} & \multicolumn{1}{l}{news., meetings} & no & \multicolumn{1}{l}{ROUGE} & no & \multicolumn{1}{l}{yes} & \multicolumn{1}{l}{informativeness, readability} & yes & no & high \\ 
40 & \citet{kikuchi2016controlling} & \multicolumn{1}{r}{2016} & EMNLP & \multicolumn{1}{l}{Length} & no & \multicolumn{1}{l}{encoder-decoder} & \multicolumn{1}{l}{ante-hoc, post-hoc} & \multicolumn{1}{l}{supervised} & yes & \multicolumn{1}{l}{{\ul \href{https://github.com/kiyukuta/lencon}{Link}}} & \multicolumn{1}{l}{DUC2004, Gigaword} & \multicolumn{1}{l}{document} & \multicolumn{1}{l}{generic} & \multicolumn{1}{l}{no} & \multicolumn{1}{l}{news} & no & \multicolumn{1}{l}{ROUGE} & no & \multicolumn{1}{l}{no} & \multicolumn{1}{l}{no} & no & no & high \\ 
41 & \citet{liu2018controlling} & \multicolumn{1}{r}{2018} & EMNLP & \multicolumn{1}{l}{Length} & no & \multicolumn{1}{l}{encoder-decoder} & \multicolumn{1}{l}{ante-hoc} & \multicolumn{1}{l}{supervised} & yes & \multicolumn{1}{l}{{\ul \href{https://github.com/YizhuLiu/sumlen}{Link}}} & \multicolumn{1}{l}{CNNDM, DMQA} & \multicolumn{1}{l}{document} & \multicolumn{1}{l}{generic} & \multicolumn{1}{l}{no} & \multicolumn{1}{l}{news} & no & \multicolumn{1}{l}{\begin{tabular}[c]{@{}l@{}}ROUGE, \\ VAR, Similarity\end{tabular}} & no & \multicolumn{1}{l}{no} & \multicolumn{1}{l}{no} & no & no & high \\ 
42 & \citet{makino-etal-2019-global} & \multicolumn{1}{r}{2019} & ACL & \multicolumn{1}{l}{Length} & no & \multicolumn{1}{l}{encoder-decoder} & \multicolumn{1}{l}{ante-hoc} & \multicolumn{1}{l}{supervised} & no & \multicolumn{1}{l}{-} & \multicolumn{1}{l}{CNNDM, Mainichi} & \multicolumn{1}{l}{document} & \multicolumn{1}{l}{generic} & \multicolumn{1}{l}{no} & \multicolumn{1}{l}{news} & no & \multicolumn{1}{l}{ROUGE} & no & \multicolumn{1}{l}{yes} & \multicolumn{1}{l}{post-editing-time} & no & no & not sure \\ 
43 & \citet{yu-etal-2021-lenatten} & \multicolumn{1}{r}{2021} & ACL & \multicolumn{1}{l}{Length} & no & \multicolumn{1}{l}{encoder-decoder} & \multicolumn{1}{l}{ante-hoc} & \multicolumn{1}{l}{supervised} & yes & \multicolumn{1}{l}{{\ul \href{https://github.com/X-AISIG/LenAtten}{Link}}} & \multicolumn{1}{l}{CNNDM} & \multicolumn{1}{l}{document} & \multicolumn{1}{l}{generic} & \multicolumn{1}{l}{no} & \multicolumn{1}{l}{news} & no & \multicolumn{1}{l}{ROUGE} & no & \multicolumn{1}{l}{yes} & \multicolumn{1}{l}{\begin{tabular}[c]{@{}l@{}}Correctness, completeness\\ fluency\end{tabular}} & no & no & high \\ 
44 & \citet{sarkhel-etal-2020-interpretable} & \multicolumn{1}{r}{2020} & COLING & \multicolumn{1}{l}{Length} & no & \multicolumn{1}{l}{encoder-decoder} & \multicolumn{1}{l}{ante-hoc} & \multicolumn{1}{l}{supervised} & no & \multicolumn{1}{l}{-} & \multicolumn{1}{l}{\begin{tabular}[c]{@{}l@{}}MSR Narratives,\\ Thinking-Machines\end{tabular}} & \multicolumn{1}{l}{document} & \multicolumn{1}{l}{generic} & \multicolumn{1}{l}{no} & \multicolumn{1}{l}{\begin{tabular}[c]{@{}l@{}}social media, \\ science/education \end{tabular}} & no & \multicolumn{1}{l}{ROUGE, METEOR} & no & \multicolumn{1}{l}{yes} & \multicolumn{1}{l}{presence of key facts} & no & no & not sure \\ 
45 & \citet{he-etal-2022-ctrlsum} & \multicolumn{1}{r}{2022} & EMNLP & \multicolumn{1}{l}{Length, Entity} & yes & \multicolumn{1}{l}{encoder-decoder} & \multicolumn{1}{l}{ante-hoc} & \multicolumn{1}{l}{supervised} & yes & \multicolumn{1}{l}{{\ul \href{https://github.com/salesforce/ctrl-sum}{Link}}} & \multicolumn{1}{l}{CNNDM, arXiv, BIGPATENT} & \multicolumn{1}{l}{document} & \multicolumn{1}{l}{generic} & \multicolumn{1}{l}{no} & \multicolumn{1}{l}{news, scientific data} & no & \multicolumn{1}{l}{ROUGE} & no & \multicolumn{1}{l}{yes} & \multicolumn{1}{l}{\begin{tabular}[c]{@{}l@{}}Control Accuracy, \\ Control Relevance\end{tabular}} & no & yes & high \\ 
46 & \citet{tan-etal-2020-summarizing} & \multicolumn{1}{r}{2020} & EMNLP & \multicolumn{1}{l}{Topic} & no & \multicolumn{1}{l}{encoder-decoder} & \multicolumn{1}{l}{ante-hoc} & \multicolumn{1}{l}{supervised} & yes & \multicolumn{1}{l}{{\ul \href{https://github.com/tanyuqian/aspect-based-summarization}{Link}}} & \multicolumn{1}{l}{CNNDM, MA News} & \multicolumn{1}{l}{document} & \multicolumn{1}{l}{generic} & \multicolumn{1}{l}{no} & \multicolumn{1}{l}{news} & no & \multicolumn{1}{l}{ROUGE} & no & \multicolumn{1}{l}{yes} & \multicolumn{1}{l}{\begin{tabular}[c]{@{}l@{}}Accuracy , Informativeness, \\ Fluency\end{tabular}} & yes & no & high \\ 
47 & \citet{narayan-etal-2022-well} & \multicolumn{1}{r}{2022} & ACL & \multicolumn{1}{l}{Diversity} & no & \multicolumn{1}{l}{encoder-decoder} & \multicolumn{1}{l}{post-hoc} & \multicolumn{1}{l}{supervised} & yes & \multicolumn{1}{l}{{\ul \href{https://github.com/google-research/language/tree/master/language/frost}{Link}}} & \multicolumn{1}{l}{\begin{tabular}[c]{@{}l@{}}CNNDM, XSum,\\ SQuAD)\end{tabular}} & \multicolumn{1}{l}{document} & \multicolumn{1}{l}{generic} & \multicolumn{1}{l}{no} & \multicolumn{1}{l}{news} & no & \multicolumn{1}{l}{\begin{tabular}[c]{@{}l@{}}ROUGE,BertScore,\\ SELF-BLEU\end{tabular}} & no & \multicolumn{1}{l}{yes} & \multicolumn{1}{l}{faithfulness and diversity} & no & no & medium \\ 
48 & \citet{zhong2023strong} & \multicolumn{1}{r}{2023} & IJCNLP-AACL & \multicolumn{1}{l}{Structure} & no & \multicolumn{1}{l}{encoder-decoder} & \multicolumn{1}{l}{ante-hoc} & \multicolumn{1}{l}{supervised} & no & \multicolumn{1}{l}{-} & \multicolumn{1}{l}{CanLII} & \multicolumn{1}{l}{document} & \multicolumn{1}{l}{generic} & \multicolumn{1}{l}{no} & \multicolumn{1}{l}{legal} & no & \multicolumn{1}{l}{\begin{tabular}[c]{@{}l@{}}ROUGE, BERTScore,\\ structure similarity.\end{tabular}} & no & \multicolumn{1}{l}{yes} & \multicolumn{1}{l}{coherence, coverage} & no & yes & not sure \\ 
49 & \citet{suhara-etal-2020-opiniondigest} & \multicolumn{1}{r}{2020} & ACL & \multicolumn{1}{l}{Topic} & yes & \multicolumn{1}{l}{extractor + generator(decoder)} & \multicolumn{1}{l}{ante-hoc} & \multicolumn{1}{l}{unsupervised} & yes & \multicolumn{1}{l}{{\ul \href{https://github.com/megagonlabs/opiniondigest}{Link}}} & \multicolumn{1}{l}{Hotel, Yelp} & \multicolumn{1}{l}{reviews} & \multicolumn{1}{l}{generic} & \multicolumn{1}{l}{no} & \multicolumn{1}{l}{reviews} & no & \multicolumn{1}{l}{ROUGE} & no & \multicolumn{1}{l}{yes} & \multicolumn{1}{l}{\begin{tabular}[c]{@{}l@{}}informativeness, coherence, \\ non-redundancy, \\ content-support\end{tabular}} & no & no & high \\ 
50 & \citet{ribeiro2023generating} & \multicolumn{1}{r}{2023} & EMNLP & \multicolumn{1}{l}{Style} & no & \multicolumn{1}{l}{encoder-decoder} & \multicolumn{1}{l}{pre-hoc, ante-hoc} & \multicolumn{1}{l}{supervised} & yes & \multicolumn{1}{l}{{\ul \href{https://github.com/amazon-science/controllable-readability-summarization}{Link}}} & \multicolumn{1}{l}{CNNDM} & \multicolumn{1}{l}{document} & \multicolumn{1}{l}{generic} & \multicolumn{1}{l}{no} & \multicolumn{1}{l}{news} & no & \multicolumn{1}{l}{\begin{tabular}[c]{@{}l@{}}ROUGE, BERTScore,\\  FRE, GFI, CLI\end{tabular}} & yes & \multicolumn{1}{l}{yes} & \multicolumn{1}{l}{most readable and least readable} & no & yes & high \\ 
51 & \citet{nallapati2017summarunner} & \multicolumn{1}{r}{2017} & AAAI & \multicolumn{1}{l}{Abstractivity, Saliency} & yes & \multicolumn{1}{l}{GRU-RNN} & \multicolumn{1}{l}{ante-hoc} & \multicolumn{1}{l}{supervised} & yes & \multicolumn{1}{l}{{\ul \href{https://github.com/hpzhao/SummaRuNNer}{Link}}} & \multicolumn{1}{l}{CNN/DM, DUC2002} & \multicolumn{1}{l}{document} & \multicolumn{1}{l}{generic} & \multicolumn{1}{l}{no} & \multicolumn{1}{l}{news} & no & \multicolumn{1}{l}{ROUGE} & no & \multicolumn{1}{l}{no} & \multicolumn{1}{l}{no} & no & no & high \\ 
52 & \citet{hitomi-etal-2019-large} & \multicolumn{1}{r}{2019} & INLG & \multicolumn{1}{l}{Length} & no & \multicolumn{1}{l}{encoder-decoder} & \multicolumn{1}{l}{pre-hoc, ante-hoc} & \multicolumn{1}{l}{supervised} & no & \multicolumn{1}{l}{-} & \multicolumn{1}{l}{JAMUS} & \multicolumn{1}{l}{document} & \multicolumn{1}{l}{generic} & \multicolumn{1}{l}{yes} & \multicolumn{1}{l}{news} & {\ul \href{https://cl.asahi.com/api\_data/jnc-jamul-en.html}{Link}} & \multicolumn{1}{l}{ROUGE} & no & \multicolumn{1}{l}{no} & \multicolumn{1}{l}{no} & no & no & medium \\ 
53 & \citet{shen2022multi} & \multicolumn{1}{r}{2022} & NIPS & \multicolumn{1}{l}{Coverage} & no & \multicolumn{1}{l}{encoder-decoder} & \multicolumn{1}{l}{pre-hoc} & \multicolumn{1}{l}{supervised} & yes & \multicolumn{1}{l}{{\ul \href{https://github.com/multilexsum/dataset}{Link}}} & \multicolumn{1}{l}{Multi-LexSum} & \multicolumn{1}{l}{document} & \multicolumn{1}{l}{Human-annotated} & \multicolumn{1}{l}{yes} & \multicolumn{1}{l}{Legal} & {\ul \href{https://multilexsum.github.io}{Link}} & \multicolumn{1}{l}{ROUGE,BERTSCORE} & no & \multicolumn{1}{l}{yes} & \multicolumn{1}{l}{Rate the generation (0-3 scale)} & no & yes & high \\ 
54 & \citet{maddela-etal-2022-entsum} & \multicolumn{1}{r}{2022} & ACL & \multicolumn{1}{l}{Entity} & no & \multicolumn{1}{l}{encoder, encoder-decoder} & \multicolumn{1}{l}{pre-hoc, ante-hoc} & \multicolumn{1}{l}{supervised} & yes & \multicolumn{1}{l}{{\ul \href{https://github.com/bloomberg/entsum}{Link}}} & \multicolumn{1}{l}{CNNDM, NYT} & \multicolumn{1}{l}{document} & \multicolumn{1}{l}{generic} & \multicolumn{1}{l}{yes} & \multicolumn{1}{l}{news} & {\ul \href{https://github.com/bloomberg/entsum}{Link}} & \multicolumn{1}{l}{ROUGE, BERTScore} & no & \multicolumn{1}{l}{no} & \multicolumn{1}{l}{no} & no & no & high \\ 
55 & \citet{bahrainian-etal-2022-newts} & \multicolumn{1}{r}{2022} & ACL & \multicolumn{1}{l}{Topic} & no & \multicolumn{1}{l}{encoder-decoder, decoder} & \multicolumn{1}{l}{-} & \multicolumn{1}{l}{supervised} & no & \multicolumn{1}{l}{-} & \multicolumn{1}{l}{NEWTS} & \multicolumn{1}{l}{document} & \multicolumn{1}{l}{Human-annotated} & \multicolumn{1}{l}{yes} & \multicolumn{1}{l}{news} & {\ul \href{https://github.com/ali-bahrainian/NEWTS}{Link}} & \multicolumn{1}{l}{ROUGE} & no & \multicolumn{1}{l}{yes} & \multicolumn{1}{l}{Binay evaluation} & yes & no & high \\ 
56 & \citet{hayashi-etal-2021-wikiasp} & \multicolumn{1}{r}{2021} & TACL & \multicolumn{1}{l}{Topic} & no & \multicolumn{1}{l}{encoder, textRank} & \multicolumn{1}{l}{pre-hoc} & \multicolumn{1}{l}{supervised} & yes & \multicolumn{1}{l}{{\ul \href{https://github.com/neulab/wikiasp}{Link}}} & \multicolumn{1}{l}{WikiAsp} & \multicolumn{1}{l}{Wikipedia} & \multicolumn{1}{l}{Automatic} & \multicolumn{1}{l}{yes} & \multicolumn{1}{l}{Wikipedia} & {\ul \href{https://huggingface.co/datasets/wiki\_asp}{Link}} & \multicolumn{1}{l}{ROUGE} & no & \multicolumn{1}{l}{yes} & \multicolumn{1}{l}{no} & no & no & high \\ 
57 & \citet{luo-etal-2022-readability} & \multicolumn{1}{r}{2022} & EMNLP & \multicolumn{1}{l}{Style} & no & \multicolumn{1}{l}{encoder-decoder} & \multicolumn{1}{l}{pre-hoc} & \multicolumn{1}{l}{supervised} & no & \multicolumn{1}{l}{-} & \multicolumn{1}{l}{TS and PLS} & \multicolumn{1}{l}{document} & \multicolumn{1}{l}{Automatic} & \multicolumn{1}{l}{yes} & \multicolumn{1}{l}{Scientific} & {\ul \href{http://www.nactem.ac.uk/readability/}{Link}} & \multicolumn{1}{l}{\begin{tabular}[c]{@{}l@{}}ROUGE, FKG, \\ CLI,ARI, RNTPC\end{tabular}} & yes & \multicolumn{1}{l}{yes} & \multicolumn{1}{l}{\begin{tabular}[c]{@{}l@{}}Relevance, Grammar, \\ Coherence\end{tabular}} & no & yes & high \\ 
58 & \citet{ahuja-etal-2022-aspectnews} & \multicolumn{1}{r}{2022} & ACL & \multicolumn{1}{l}{Topic} & no & \multicolumn{1}{l}{encoder, encoder-decoder} & \multicolumn{1}{l}{pre-hoc} & \multicolumn{1}{l}{supervised} & yes & \multicolumn{1}{l}{{\ul \href{https://github.com/tanyuqian/aspect-based-summarization}{Link}}} & \multicolumn{1}{l}{ASPECTNEWS} & \multicolumn{1}{l}{document} & \multicolumn{1}{l}{Automatic} & \multicolumn{1}{l}{yes} & \multicolumn{1}{l}{news} & {\ul \href{https://github.com/oja/aosumm/tree/master/data}{Link}} & \multicolumn{1}{l}{ROUGE} & no & \multicolumn{1}{l}{no} & \multicolumn{1}{l}{no} & no & no & Medium \\ 
59 & \citet{mukherjee2020read} & \multicolumn{1}{r}{2020} & SIGIR & \multicolumn{1}{l}{Topic} & no & \multicolumn{1}{l}{Graph-based} & \multicolumn{1}{l}{ante-hoc} & \multicolumn{1}{l}{Unsupervised} & yes & \multicolumn{1}{l}{{\ul \href{https://github.com/rajdeep345/ControllableSumm}{Link}}} & \multicolumn{1}{l}{Tourism Reviews} & \multicolumn{1}{l}{document} & \multicolumn{1}{l}{Automatic} & \multicolumn{1}{l}{yes} & \multicolumn{1}{l}{Tourism} & {\ul \href{https://github.com/rajdeep345/ControllableSumm}{Link}} & \multicolumn{1}{l}{ROUGE} & no & \multicolumn{1}{l}{yes} & \multicolumn{1}{l}{\begin{tabular}[c]{@{}l@{}}aspect-coverage, readability,\\ diversity\end{tabular}} & no & no & high \\ 
60 & \citet{lin-etal-2021-csds} & \multicolumn{1}{r}{2021} & EMNLP & \multicolumn{1}{l}{Role} & no & \multicolumn{1}{l}{all} & \multicolumn{1}{l}{pre-hoc} & \multicolumn{1}{l}{supervised} & yes & \multicolumn{1}{l}{{\ul \href{https://github.com/xiaolinAndy/CSDS}{Link}}} & \multicolumn{1}{l}{CSDS} & \multicolumn{1}{l}{reviews} & \multicolumn{1}{l}{Human-annotated} & \multicolumn{1}{l}{yes} & \multicolumn{1}{l}{reviews} & {\ul \href{https://github.com/Shen-Chenhui/MReD}{Link}} & \multicolumn{1}{l}{ROUGE, BERTScore} & no & \multicolumn{1}{l}{yes} & \multicolumn{1}{l}{\begin{tabular}[c]{@{}l@{}}Informativeness, \\ non-redundancy, \\ fluency, matching rate\end{tabular}} & yes & no & high \\ 
61 & \citet{shen-etal-2022-mred} & \multicolumn{1}{r}{2022} & ACL & \multicolumn{1}{l}{Structure} & no & \multicolumn{1}{l}{encoder-decoder} & \multicolumn{1}{l}{pre-hoc} & \multicolumn{1}{l}{supervised} & yes & \multicolumn{1}{l}{{\ul \href{https://github.com/Shen-Chenhui/MReD}{Link}}} & \multicolumn{1}{l}{MReD} & \multicolumn{1}{l}{reviews} & \multicolumn{1}{l}{Human-annotated} & \multicolumn{1}{l}{yes} & \multicolumn{1}{l}{Scientific} & {\ul \href{https://github.com/Shen-Chenhui/MReD}{Link}} & \multicolumn{1}{l}{ROUGE} & no & \multicolumn{1}{l}{yes} & \multicolumn{1}{l}{\begin{tabular}[c]{@{}l@{}}Fluency, content relevance,  \\ structure, decision correctness\end{tabular}} & no & no & high \\ \bottomrule
\end{tabular}%
}
\caption{Master survey table (*marked fields are filled to best of our understanding based on the available information).}
\label{tab:checklist}
\end{sidewaystable}
\end{document}